\documentclass[10pt,twocolumn,letterpaper]{article}

\usepackage[pagenumbers]{cvpr} % To force page numbers, e.g. for an arXiv version

\newcommand{\settablefont}{\fontsize{8}{12}\selectfont}

\definecolor{roadcolor}{RGB}{128,64,128}
\definecolor{sidewalkcolor}{RGB}{244,35,232}
\definecolor{buildingcolor}{RGB}{70,70,70}
\definecolor{wallcolor}{RGB}{102,102,156}
\definecolor{fencecolor}{RGB}{190,153,153}
\definecolor{polecolor}{RGB}{153,153,153}
\definecolor{trafficlightcolor}{RGB}{250,170,30}
\definecolor{trafficsigncolor}{RGB}{220,220,0}
\definecolor{vegetationcolor}{RGB}{107,142,35}
\definecolor{groundcolor}{RGB}{152,251,152}
\definecolor{personcolor}{RGB}{220,20,60}
\definecolor{carcolor}{RGB}{0,0,142}
\definecolor{truckcolor}{RGB}{0,0,70}
\definecolor{othervehiclecolor}{RGB}{0,60,100}
\definecolor{othercolor}{RGB}{70,130,180}

\usepackage{cuted}
\usepackage{mathrsfs}
\usepackage{multirow}
\usepackage{makecell}
\usepackage{bbm}

\definecolor{cvprblue}{rgb}{0.21,0.49,0.74}
\usepackage[pagebackref,breaklinks,colorlinks,allcolors=cvprblue]{hyperref}

\def\paperID{6543} % *** Enter the Paper ID here
\def\confName{CVPR}
\def\confYear{2026}

\title{An Instance-Centric Panoptic Occupancy Prediction Benchmark \\ for Autonomous Driving}

\author{
	Yi Feng\textsuperscript{1*}\\
	\and
	Junwu E\textsuperscript{1*}\\
	\and
	Zizhan Guo\textsuperscript{1}\\
	\and
	Yu Ma\textsuperscript{1}\\
	\and
	Hanli Wang\textsuperscript{1}\\
	\and
	Rui Fan\textsuperscript{1}\textsuperscript{$\dagger$}\\
	\and 
	\textsuperscript{1}Tongji University \\
}

\begin{document}
\maketitle

\begin{strip}
	\vspace{-3em}
	\centering
	\includegraphics[width=\textwidth]{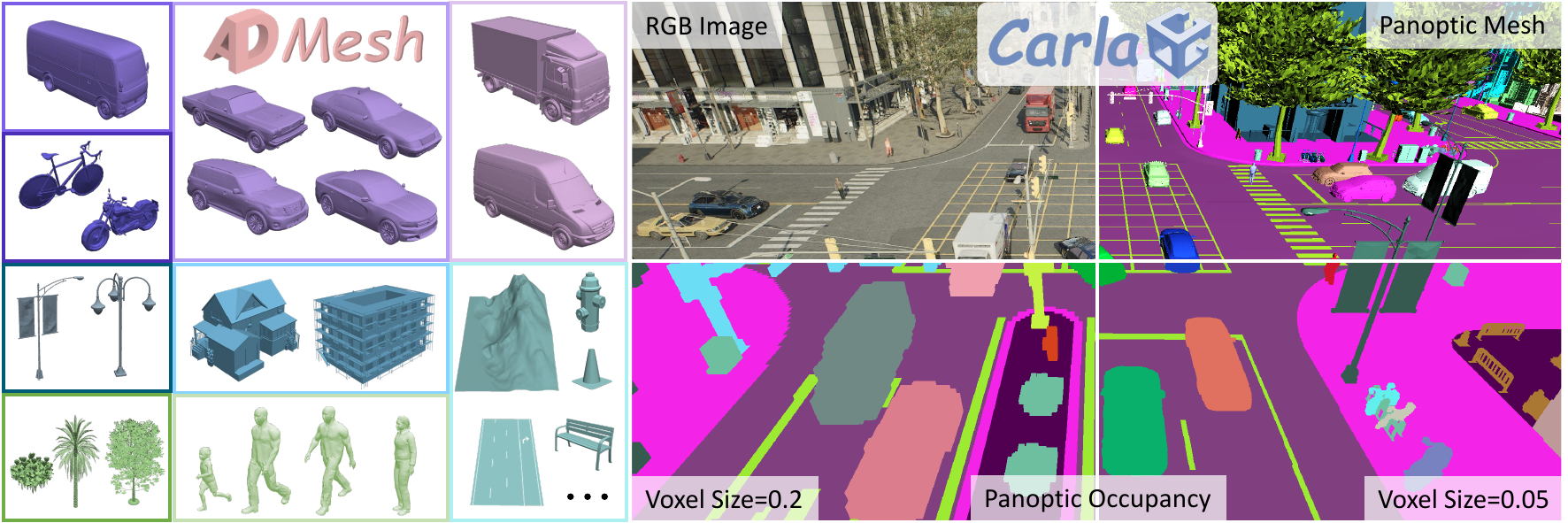}
	\captionof{figure}{Overview of the proposed benchmark. \textbf{ADMesh} provides the first large-scale, semantically structured 3D mesh library for autonomous driving. \textbf{CarlaOcc} leverages these assets to construct a multi-modal, high-fidelity, and physically consistent panoptic occupancy dataset, featuring variable voxel resolutions and rich instance-level annotations for comprehensive 3D perception benchmarking.}
	\label{fig.teaser}
\end{strip}

\begingroup
\renewcommand\thefootnote{}
\footnotetext{\textsuperscript{*}Equal Contribution.\ \ \ \ \ \textsuperscript{$\dagger$}Corresponding author.}
\endgroup

\begin{abstract}
	Panoptic occupancy prediction aims to jointly infer voxel-wise semantics and instance identities within a unified 3D scene representation. Nevertheless, progress in this field remains constrained by the absence of high-quality 3D mesh resources, instance-level annotations, and physically consistent occupancy datasets. Existing benchmarks typically provide incomplete and low-resolution geometry without instance-level annotations, limiting the development of models capable of achieving precise geometric reconstruction, reliable occlusion reasoning, and holistic 3D understanding. To address these challenges, this paper presents an instance-centric benchmark for the 3D panoptic occupancy prediction task. Specifically, we introduce ADMesh, the first unified 3D mesh library tailored for autonomous driving, which integrates over 15K high-quality 3D models with diverse textures and rich semantic annotations. Building upon ADMesh, we further construct CarlaOcc, a large-scale, physically consistent panoptic occupancy dataset generated using the CARLA simulator. This dataset contains over 100K frames with fine-grained, instance-level occupancy ground truth at voxel resolutions as fine as 0.05\,m. Furthermore, standardized evaluation metrics are introduced to quantify the quality of existing occupancy datasets. Finally, a systematic benchmark of representative models is established on the proposed dataset, which provides a unified platform for fair comparison and reproducible research in the field of 3D panoptic perception. Code and dataset are available at \url{https://mias.group/CarlaOcc}.
\end{abstract}

% =============== Introduction ========================
\section{Introduction}
As a fundamental component of autonomous driving, 3D scene perception empowers a holistic understanding of surrounding environments, recognition of objects, and reasoning about their spatial relationships for safe and reliable navigation~\cite{uniscene2025Li, visionpad2025zhang, vipocc2025feng, feng2024scipad, dcpidepth2025zhang}.
Recent advances in 3D perception have evolved from purely semantic occupancy prediction toward fine-grained, instance-aware scene understanding. This shift has led to the task of \emph{Panoptic Occupancy Prediction}, which unifies semantic and instance understanding with a comprehensive and temporally consistent scene representation~\cite{sparseocc2024liu}.

However, research in this direction has been substantially constrained by the absence of instance labels in existing datasets~\cite{occ3d2023tian, surroundocc2023wei, sscbench2024li, collaborative2024song}. As a temporary workaround, recent methods such as SparseOcc~\cite{sparseocc2024liu} and PaSCo~\cite{pasco2024cao} attempt to generate pseudo panoptic labels by grouping voxels inside the 3D bounding boxes or clustering semantic occupancy grids based on existing semantic occupancy benchmarks.
Nevertheless, such heuristics inevitably introduce boundary artifacts, overlapping instances, and incomplete geometry. More critically, the quality of these pseudo labels is inherently constrained by the imperfections of the source datasets: so far, public 3D occupancy prediction datasets~\cite{occ3d2023tian, sscbench2024li} are generally limited by low and fixed resolution as well as inconsistent and fragmented geometry. This is primarily caused by their ground truth generation pipelines: both real-world datasets (\textit{e.g.}, Occ3D~\cite{occ3d2023tian}, SurroundOcc~\cite{surroundocc2023wei}) and synthetic datasets (\textit{e.g.}, CoHFF~\cite{collaborative2024song}, CarlaSC~\cite{motionsc2022wilson}) typically aggregate LiDAR scans followed by voxelization or surface reconstruction~\cite{poisson2006kazhdan}.
Due to the rapid decay of LiDAR point density with distance, these pipelines must adopt coarse, fixed voxel resolutions to balance ground truth quality between near and far ranges.
Moreover, this process encodes only the surfaces observed by sensors and cannot recover unobserved geometry, inevitably yielding incomplete scenes with missing thin, transparent structures and occlusion-induced cavities. Models trained under such supervision tend to tolerate, or even replicate these artifacts, ultimately degrading spatial understanding and occlusion reasoning in downstream tasks.

Constructing a physically consistent panoptic occupancy dataset fundamentally relies on the availability of high-quality 3D model assets. However, there remains an absence of a comprehensive 3D model library tailored for autonomous driving.
Existing 3D object datasets~\cite{omniobject3d2023wu, buildingnet2021selvaraju, 3drealcar2025du, objaverse2023deitke} provide high-quality static meshes, yet they are either generic or narrowly scoped to a single category, lacking hierarchical semantics that align with autonomous driving taxonomies. Simulation engines such as CARLA~\cite{carla2017dosovitskiy} and Waymax~\cite{waymax2023gulino} offer driving-related content but are fragmented and platform-dependent, limiting their reusability for large-scale, high-fidelity dataset construction.

These limitations motivate us to construct a unified, comprehensive, and instance-centric framework, which provides structured, high-quality mesh resources, physically consistent occupancy annotations, and a standardized benchmark for 3D panoptic perception.
As shown in Fig.~\ref{fig.teaser}, this paper introduces \textbf{ADMesh}, the \emph{first} unified 3D mesh library with hierarchically organized semantic taxonomy tailored for autonomous driving. Derived from four established open-source resources, including the CARLA simulation platform~\cite{carla2017dosovitskiy}, BuildingNet~\cite{buildingnet2021selvaraju}, MeshFleet~\cite{meshfleet2025boborzi}, and ShapeNet~\cite{shapenet2015chang}, ADMesh integrates over 15K high-quality 3D models with diverse textures and rich annotations, providing a comprehensive, high-fidelity resource for large-scale scene reconstruction, controllable scene generation, and instance-centric perception research.
Building upon ADMesh, we further construct \textbf{CarlaOcc}, a large-scale and physically consistent panoptic occupancy dataset generated with the CARLA simulator~\cite{carla2017dosovitskiy} powered by Unreal Engine~5~\cite{unrealengine5}. CarlaOcc contains over 100K training frames across diverse driving scenes, featuring complex urban layouts, dynamic traffic agents, and high-fidelity ground-truth labels.
Unlike previous datasets that are limited by coarse voxel resolution and the absence of instance information, CarlaOcc leverages the 3D assets in ADMesh to perform scene reconstruction at \emph{arbitrary} voxel resolutions, yielding physically accurate panoptic occupancy annotations and instance-level traffic information.
To quantitatively assess the realism and internal consistency of existing occupancy datasets, we further propose two novel metrics that measure geometric continuity, semantic completeness, and temporal consistency in occupancy labeling. 
We further benchmark representative models across diverse 3D perception tasks, providing the community with a standardized platform for fair comparison and reproducible evaluation.

% =============== Related Work ========================
\section{Related Work}
\noindent\textbf{Panoptic Occupancy Prediction.}
Panoptic segmentation was first introduced to provide a holistic understanding of a 2D image by combining the goals of both semantic and instance segmentation~\cite{panoptic2019kirillov}.
Extending this concept to the 3D domain, LiDAR-based panoptic segmentation has progressed toward end-to-end, proposal-free, and clustering-free frameworks~\cite{masklidar2023marcuzzi, tpvformer2023huang}.
Beyond labeling only the observed points, panoptic scene completion further infers the occluded geometry and assigns panoptic labels in 3D. As the first work to formalize this task, PaSCo~\cite{pasco2024cao} predicts a complete panoptic scene with incorporated uncertainty information, facilitating a more comprehensive understanding of the scene.
In vision-centric perception, panoptic occupancy prediction extends semantic occupancy by jointly estimating voxel-wise semantics and instance identities in 3D space.
SparseOcc~\cite{sparseocc2024liu} introduces a fully sparse pipeline that reconstructs sparse 3D representations and predicts both semantic and instance occupancy by leveraging sparse queries in a mask Transformer. Concurrently, PanoOcc~\cite{panoocc2024wang} proposes a unified occupancy representation for camera-based 3D panoptic segmentation, using voxel queries to aggregate multi-view information in a coarse-to-fine scheme. Panoptic-FlashOcc~\cite{flashocc2024yu} presents an efficient baseline that couples semantic occupancy with panoptic prediction with instance centers, achieving a favorable balance between speed and accuracy.
Despite promising progress, public panoptic datasets remain limited, and the lack of standardized, instance-centric mesh resources continues to constrain research on unified panoptic 3D perception.

\noindent\textbf{3D Object Datasets.}
A variety of large-scale 3D object datasets have been developed to support geometric modeling and scene understanding. Early works such as ShapeNet~\cite{shapenet2015chang} and ABO~\cite{abo2022collins} provided generic object-level CAD models covering a wide range of categories. More recent efforts, including OmniObject3D~\cite{omniobject3d2023wu} and Objaverse-XL~\cite{objaverse2023deitke}, provide high-quality multi-view meshes that support research on 3D perception, geometry reconstruction, and generative scene modeling. However, these datasets are primarily designed for isolated or indoor objects and lack hierarchical semantics or taxonomies consistent with autonomous driving, limiting their applicability to large-scale outdoor environments and traffic-centric 3D scene understanding.
Vehicle-focused datasets, such as 3DRealCar~\cite{3drealcar2025du}, MeshFleet~\cite{meshfleet2025boborzi}, and DrivAerNet++~\cite{drivaernet++2024elrefaie}, obtain high-quality automotive models through real-world scanning, professional design, and aerodynamic simulation, respectively, while BuildingNet~\cite{buildingnet2021selvaraju} focuses on detailed architectural structures with rich part-level annotations. 
Despite their domain-specific precision, these datasets remain narrowly scoped and lack the hierarchical semantic organization required for scalable autonomous driving research.
Meanwhile, large-scale simulation platforms, including CARLA~\cite{carla2017dosovitskiy}, Waymax~\cite{waymax2023gulino}, and LGSVL~\cite{lgsvl2020rong}, provide diverse virtual assets for driving simulation, yet these resources are fragmented, non-standardized, and platform-dependent.
Overall, there remains a remarkable gap for a semantically structured, standardized 3D mesh database tailored for autonomous driving, providing unified taxonomy, consistent semantics, and comprehensive coverage of vehicles, infrastructure, and environmental elements for scalable simulation and perception research.

\noindent\textbf{Occupancy Prediction Datasets and Benchmarks.}
Autonomous driving perception has been significantly advanced by large-scale datasets, with the pioneering KITTI suite~\cite{kitti2012geiger, kitti2013geiger, objectsceneflow2015menze} establishing the foundation for 3D vision in driving. Subsequent benchmarks, such as KITTI-360~\cite{kitti360_2022liao}, nuScenes~\cite{nuscenes2020caesar}, and Waymo~\cite{waymo2020sun}, further expand scale and sensor diversity, providing synchronized LiDAR-camera data with dense 3D annotations for detection, tracking, and mapping. However, these benchmarks focus primarily on sparse point-level supervision and do not provide dense 3D occupancy labels.
To address this limitation, a series of occupancy prediction benchmarks has emerged.
Early works, such as SemanticKITTI~\cite{semantickitti2019behley}, SemanticPOSS~\cite{semanticposs2020pan}, and SynthCity~\cite{synthcity2020pham}, extended LiDAR data with semantic occupancy annotations.
More recent datasets such as Occ3D~\cite{occ3d2023tian}, SurroundOcc~\cite{surroundocc2023wei}, OpenOccupancy~\cite{openoccupancy2023wang}, and SSCBench~\cite{sscbench2024li} established unified evaluation protocols and provided larger-scale outdoor coverage, significantly advancing semantic occupancy research.
Nevertheless, their ground truth generation pipelines rely on voxelization or surface reconstruction from aggregated LiDAR scans, which capture only sensor-visible regions.
As a result, these datasets often exhibit low and fixed resolution, incomplete geometry, and physically inconsistent structures.
Synthetic datasets such as CarlaSC~\cite{motionsc2022wilson} and CoHFF~\cite{collaborative2024song} leverage the CARLA simulator~\cite{carla2017dosovitskiy} to significantly enhance both the quality and scope of ground truth annotations, but remain limited by the realism and diversity of simulation assets.
Overall, existing occupancy benchmarks focus primarily on \emph{semantic} occupancy without explicit instance information, leaving panoptic reasoning, instance-level continuity, and temporally consistent 3D scene understanding largely unexplored.

% =============== Methodology ========================
\section{The Panoptic Occupancy Benchmark}

\begin{table}[!t]
	\centering
	\settablefont
	\caption{Overview of data sources and statistics of ADMesh.}
	\label{tab.ADMesh}
	\begin{tabular}{lll}
		\toprule[1pt]
		\textbf{Source} & \textbf{Instances} & \textbf{Main Categories}  \\
		\midrule
		CARLA~\cite{carla2017dosovitskiy} & $\sim$8,000 & 30 Standardized Categories\\
		BuildingNet~\cite{buildingnet2021selvaraju} & $\sim$2,000 & Building \\
		MeshFleet~\cite{meshfleet2025boborzi} & $\sim$3,000 & Car, Truck, Bus  \\
		ShapeNetCore~\cite{shapenet2015chang} & $\sim$2,000 & Car, Train, Bus, Motorcycle  \\
		\midrule
		\textbf{Total} & \textbf{15,000+} & 30 Standardized Categories \\
		\bottomrule[1pt]
	\end{tabular}
%	\vspace{-1.0 em}
\end{table}

\begin{figure*}[!t]
	\centering
	\includegraphics[width=0.99\textwidth]{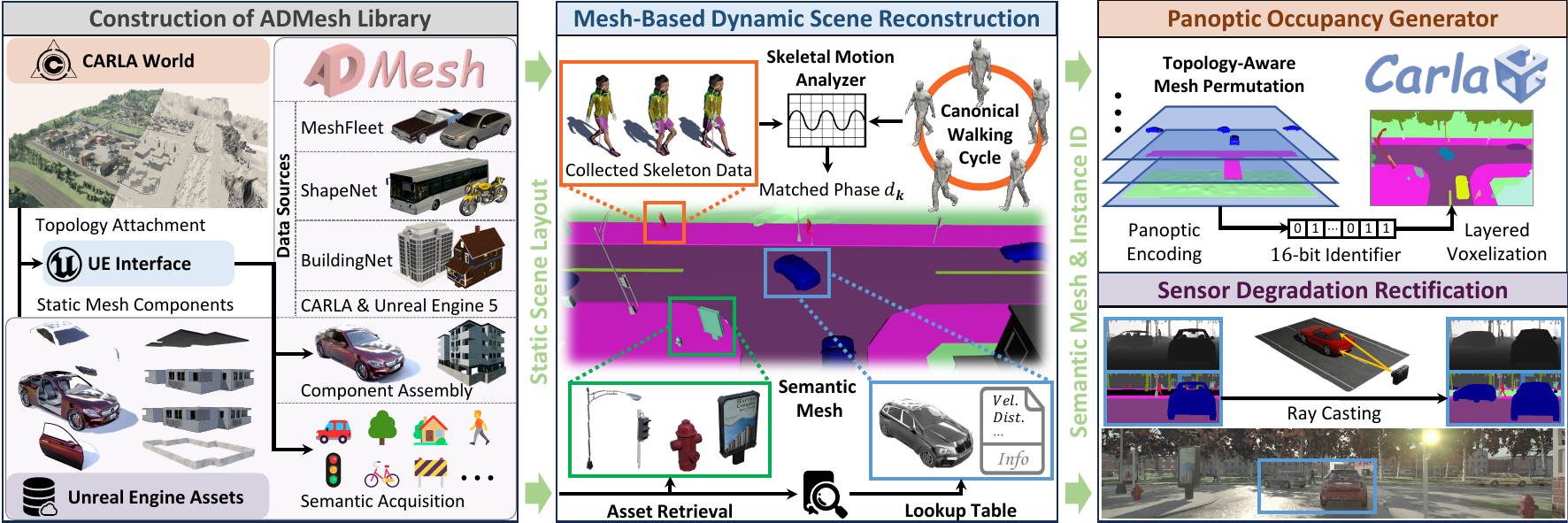}
	\caption{Overview of the proposed ADMesh library and the CarlaOcc generation pipeline. 
	The ADMesh library is constructed by extracting and organizing diverse 3D assets from multiple sources, which are subsequently used to reconstruct dynamic scenes with both static structures and temporally aligned non-rigid motions. The resulting unified scene meshes are then used to rectify sensor artifacts and further processed with a topology-aware mesh permutation strategy to produce non-overlapping panoptic occupancy labels.}
	\label{fig.method}
%	\vspace{-1.0 em}
\end{figure*}

\subsection{ADMesh: A Large-Scale Static Mesh Library}
High-quality 3D geometry is essential for autonomous driving research, yet existing mesh datasets are often limited to indoor environments or restricted in scope. To bridge this gap, we present ADMesh, a large-scale, semantically structured 3D mesh library designed specifically for autonomous driving applications. 
First, we propose a Mesh Exportation Toolchain to export the rich virtual assets and semantic annotations provided by the CARLA simulator~\cite{carla2017dosovitskiy}.
To ensure scalability and diversity, a unified framework is developed to standardize geometric representations and semantic organization, which enables seamless integration of additional resources, including BuildingNet~\cite{buildingnet2021selvaraju}, MeshFleet~\cite{meshfleet2025boborzi}, and ShapeNetCore~\cite{shapenet2015chang}. 
Through this unified framework, ADMesh includes over 15K high-quality meshes with consistent taxonomy, and effectively supports large-scale 3D scene modeling and perception research.

\noindent\textbf{Mesh Exportation Toolchain.} Virtual simulation assets, while abundant, pose several challenges for structured dataset construction:  (1) instances in virtual scenes exhibit significant redundancy and complexity; (2) individual objects are typically composed of multiple sub-meshes rather than existing as self-contained entities; (3) semantic labels are typically absent from static assets and are instead produced by the runtime rendering, and (4) asset formats vary across platforms, complicating standardized processing.

To address these challenges, we develop an automated mesh exportation toolchain, which traverses all default CARLA scenes and systematically extracts, reconstructs, and organizes 3D assets from simulation environments. 
As shown in Fig.~\ref{fig.method}, the workflow begins by exporting all valid static mesh assets, thereby creating a comprehensive component-level repository that serves as the foundational building blocks for the reconstruction of 3D models of scene objects. 
Next, we retrieve the static mesh components associated with each instantiated scene object by querying its component hierarchy through Unreal Engine's editor scripting interface, and record their local transforms as well as their attachment topology within the object's component tree. 
To ensure semantic consistency with CARLA's taxonomy, we integrate a semantic labeling module that hooks into the simulator's native annotation system, automatically mapping each scene object to its corresponding semantic category. 
Finally, leveraging the component transforms, relative attachments, semantic annotations, and the previously exported mesh assets, we hierarchically assemble each scene object from the retrieved components and reconstruct the complete object-level mesh with consistent geometry and semantics.

\noindent\textbf{Library Structure and Statistics.}
To enhance compatibility and scalability, a standardized data organization framework is developed to unify heterogeneous assets from multiple sources under a consistent representation. The framework ensures uniform naming conventions, coordinate systems, and semantic hierarchy, enabling seamless integration, indexing, and retrieval across the large-scale database. 
Within this framework, semantics are organized under a hierarchy of 30 top-level categories, further refined into fine-grained sub-classes that span the full range of driving scene elements.
Each asset ships with a standardized mesh file, associated material and texture assets, and a structured metadata record which captures the physical properties (\textit{e.g.}, semantic identifiers, geometric statistics) required by downstream pipelines. 
As shown in Table~\ref{tab.ADMesh}, ADMesh incorporates 3D mesh resources from BuildingNet~\cite{buildingnet2021selvaraju}, MeshFleet~\cite{meshfleet2025boborzi}, and ShapeNetCore~\cite{shapenet2015chang}, aggregating more than 15{,}000 high-quality models, which lays the groundwork for controllable scene generation, interactive scene editing, and geometry-aware scene perception.

% =============== Dataset Comparison ========================
\begin{table*}[!t]
	\centering
	\settablefont
	\caption{
		Comparison between CarlaOcc and other public occupancy prediction datasets.
	}
	\label{tab.dataset_comparison}
	\begin{tabular}{l|cccllcc}
		\toprule[1pt]
		Dataset & Synthetic & Surround-View & Classes & Frames & Volume Range (m) & Voxel Size (m) & Instance Annotations \\
		\midrule
		SemanticKITTI~\cite{semantickitti2019behley} 	& $\times$ 		& $\times$ 		& 28 & 43,000 	& [51.2, 51.2, 6.4] & 0.2 & $\times$ \\
		KITTI-360-SSCBench~\cite{sscbench2024li} 		& $\times$ 		& $\times$ 		& 19 & 12,865 	& [51.2, 51.2, 6.4] & 0.2 & $\times$ \\
		Occ3D-nuScenes~\cite{occ3d2023tian} 			& $\times$ 		& $\checkmark$ 	& 17 & 34,149 	& [80, 80, 6.4] 	& 0.4 & $\times$ \\
		SurroundOcc~\cite{surroundocc2023wei} 			& $\times$ 		& $\checkmark$ 	& 17 & 34,149 	& [100, 100, 8] 	& 0.5 & $\times$ \\
		CoHFF~\cite{collaborative2024song} 				& $\checkmark$ 	& $\checkmark$ 	& 12 & 11,464 	& [40, 40, 3.2] 	& 0.4 & $\times$ \\
		CarlaSC~\cite{motionsc2022wilson} 				& $\checkmark$ 	& $\checkmark$ 	& 11 & 43,200 	& [51.2, 51.2, 3] 	& 0.4 & $\times$ \\
		\midrule
		\textbf{CarlaOcc (Ours)} & $\checkmark$ & $\checkmark$ & \textbf{30} & \textbf{104,000} & \textbf{[76.8, 51.2, 13]} & \textbf{0.05} & $\checkmark$ \\
		\bottomrule[1pt]
	\end{tabular}
%	\vspace{-1.0 em}
\end{table*}

\subsection{The CarlaOcc Dataset}
\noindent\textbf{Dataset Statistics.} We generate a large-scale, multi-modal, object-centric 3D occupancy dataset using the CARLA simulator~\cite{carla2017dosovitskiy} and the photorealistic rendering engine Unreal Engine~5~\cite{unrealengine5}. The dataset contains 104 distinct driving sequences in eight synthetic towns, with each town providing thousands of high-fidelity background meshes and instance-level annotations derived from ADMesh. Each sequence consists of 1,000 temporally consistent frames with synchronized RGB images, LiDAR point clouds, and sensor poses, as well as rich annotations including detailed metadata of surrounding traffic participants and rectified semantic and depth maps.

CarlaOcc provides high-quality panoptic occupancy ground truth generated entirely from 3D meshes rather than discrete point clouds, associating each occupied voxel with both semantic and instance identities. In total, the dataset contains over 40K individual object instances across 30 standardized semantic categories aligned with the CARLA taxonomy, and also provides semantic mapping tools for compatibility with taxonomies of other datasets such as nuScenes~\cite{nuscenes2020caesar} and KITTI-360~\cite{kitti360_2022liao}.
Meanwhile, the mesh-based generation pipeline also 
supports voxelization at resolutions as fine as 0.05\,m, which is four times finer than the most detailed existing dataset, SemanticKITTI~\cite{semantickitti2019behley}. To support both single-view and surround-view occupancy prediction settings, the occupancy region is defined with respect to the LiDAR center, covering 51.2\,m forward, 25.6\,m backward, 25.6\,m to each side, and 13\,m in height.
As summarized in Table~\ref{tab.dataset_comparison}, CarlaOcc surpasses existing public occupancy benchmarks in terms of scale, perception range, and spatial resolution, while introducing the \emph{first} instance-level occupancy annotations in this domain.

\noindent\textbf{Data Collection.} 
To facilitate seamless adaptation of existing perception methods, we adopt the same sensor configuration from KITTI-360~\cite{kitti360_2022liao}, replicating its camera-LiDAR setup and extending it with additional rear-view cameras to achieve full 360$^\circ$ perception. The sensor suite comprises six surround-view cameras capturing RGB, depth, and semantic images at a resolution of 1,408$\times$376 pixels, and a 32-channel LiDAR with 80\,m sensing range and 250K points/s sampling rate. All sensors are rigidly mounted on the ego vehicle, temporally synchronized at 20\,Hz. 

To increase scene diversity, we collect 104 distinct driving sequences across eight meticulously refined CARLA towns, covering a broad range of architectural layouts and traffic infrastructure. Each town is simulated under four traffic complexity levels, which is achieved by varying vehicle density, pedestrian flow, and behavioral parameters. We record fine-grained object-level annotations for all traffic participants within an 80\,m radius, including 3D bounding boxes, velocities, and instance identities, as well as skeletal keypoints for pedestrians to support motion analysis and dynamic occupancy generation. In total, the dataset comprises over 100K multi-modal frames with synchronized sensor data and comprehensive instance-level annotations.

\noindent\textbf{Mesh-Based Scene Reconstruction.} To ensure physically consistent and geometrically complete scene representations, we reconstruct each simulation frame directly from exported static meshes and dynamic animation sequences, rather than relying on sparse LiDAR point samples. 
Let $\mathcal{S}=\{(\mathcal{M}_i, \mathbf{T}_i)\}_{i=1}^{N}$ be the set of all valid background meshes $\mathcal{M}_i$ in the simulation world, each associated with a rigid-body transformation $\mathbf{T}_i\!\in\!SE(3)$ that maps the mesh from local to world coordinates.
Given the LiDAR-anchored occupancy range $\mathcal{R}$ and its transformation $\mathbf{T}_l$ at current simulation frame, we first select the subset of background meshes that intersect with the occupancy region:
\begin{equation}
	\mathcal{S}_{\mathrm{bg}}
	=\big\{(\mathcal{M}_i,\mathbf{T}_i)\in\mathcal{S}\ \big|\ 
	\mathbf{T}_l^{-1}\mathbf{T}_i(\mathcal{M}_i)\ \cap\ \mathcal{R}\neq\varnothing\big\}.
\end{equation}
To properly reconstruct the foreground scene objects, we retrieve all rigid entities $\mathcal{A}^r=\{a_j \mid a_j \cap \mathcal{R} \neq \varnothing\}$ located within $\mathcal{R}$, recording their simulator-dependent identifiers and world transformations $\mathbf{T}_j$. The rigid foreground instance set is then constructed as $\mathcal{S}_{\mathrm{fg}}^{\mathrm{r}}=\big\{(\mathcal{M}_j,\mathbf{T}_j)\ \big|\ \mathcal{M}_j=\mathrm{LUT}(a_j), a_j\in\mathcal{A}^r \big\}$, where $\mathrm{LUT}(\cdot)$ denotes the structured lookup table that maps each identifier to its corresponding template mesh provided by ADMesh.

To faithfully reconstruct non-rigid scene objects such as pedestrians, whose surface geometry continuously deforms with skeletal motion, we introduce a \emph{Skeletal Motion Analyzer}. 
Specifically, we preprocess the walking animation assets into a canonical gait cycle consisting of $D$ discrete phases, and store the posed meshes $\{\mathcal{M}_d\}_{d=1}^{D}$ together with their bone transformations as template keyframes. 
Each phase $d$ is associated with a gait descriptor $\delta_d$ that encodes its characteristic motion pattern.  
During simulation, for the $k$-th pedestrian, we record its global transformation $\mathbf{T}_k$ and bone transformations, and further analyze the skeletal motion trajectory within a sliding window to obtain its gait descriptor $\delta_k$. 
The instantaneous gait phase is then estimated by aligning this descriptor to the canonical walking cycle via geodesic matching:
\begin{equation}
	d_k = \arg\min_{d}\ \mathcal{G} (\delta_k,\delta_d),
\end{equation}
where $\mathcal{G}$ denotes the geodesic discrepancy measurement. By pairing the matched template mesh with global transformation, the non-rigid foreground mesh set can be represented as $\mathcal{S}_{\mathrm{fg}}^{\mathrm{n}}=\big\{(\mathcal{M}_{d_k}, \mathbf{T}_k)\}$, 
and the frame-level panoptic scene mesh is constructed by uniting background, rigid, and non-rigid foreground meshes in the world coordinates:
\begin{equation}
	\mathcal{M}^{\mathrm{pano}}
	=\big\{\,\big(\mathbf{T}(\mathcal{M}),\,p\big)\ \big|\ 
	(\mathcal{M},\mathbf{T})\in
	\mathcal{S}_{\mathrm{bg}}\cup
	\mathcal{S}_{\mathrm{fg}}^{\mathrm{r}}\cup
	\mathcal{S}_{\mathrm{fg}}^{\mathrm{n}}
	\,\big\},
\end{equation}
where $p$ denotes the panoptic label of mesh $\mathcal{M}$ that stores both semantic label and instance ID. The assembled panoptic scene mesh $\mathcal{M}^{\mathrm{pano}}$ provides a unified 3D representation that preserves both geometric fidelity and semantic completeness, serving as the foundation for subsequent voxelization and panoptic occupancy generation.

\noindent\textbf{Panoptic Occupancy Generation.}
A straightforward approach to generate panoptic occupancy labels is to independently voxelize each scene object mesh and then merge them into a global occupancy grid. 
However, such per-mesh voxelization is computationally prohibitive for large scenes, and instance-level composition leads to label conflicts in spatially overlapping regions. 
To overcome these limitations, we introduce a topology-aware mesh permutation strategy, which produces physically consistent and geometrically continuous panoptic occupancy labels using the unified panoptic scene mesh. 
We first group all stuff meshes by their semantic categories and fuse them within the predefined occupancy region, effectively eliminating redundant boundaries and significantly improving computational efficiency. Subsequently, all instances and stuff meshes are sorted according to their world-space elevation. Starting from the lowest layer, each instance mesh is discretized and integrated into the occupancy grid in ascending order of height, which ensures the lower structures (\textit{e.g.}, terrain) do not overwrite or occlude higher ones (\textit{e.g.}, road).

\noindent\textbf{Instance-Guided Rectification of Sensor Artifacts. }
In practice, the collected multi-modal data inevitably contain artifacts caused by sensor degradation and rendering failures. 
As shown in Fig.~\ref{fig.method}, it can be frequently observed that for all transparent or semi-transparent objects, both the depth and semantic maps are incorrectly marked as those of the opaque objects behind them. Moreover, certain instances completely lose their categorical labels and are instead labeled with the semantics of background objects in the semantic maps. These errors result in significant cross-modal inconsistencies, where instance boundaries, depth discontinuities, and semantic regions no longer align, thus corrupting the physical integrity of the dataset.

To address these issues, we introduce an instance-guided post-processing pipeline that leverages instance-level cues to identify and correct these degraded regions, without modifying the underlying rendering engine. 
The pipeline works by first creating a scene mesh that contains only the geometries of transparent or semi-transparent objects, as well as those exhibiting semantic inconsistencies. Ray casting is then adopted to generate accurate depth maps by tracing rays through the scene and recording the first visible intersections with the semantic meshes. 
These ray-casted depth maps are then integrated with the raw sensor measurements via pointwise minimum selection, thereby correcting depth errors caused by transparency and occlusions. For the semantic map, instance-level labels are refined by using ray-casted geometry to correct mislabeling, especially in cases where transparent objects cause semantic misclassifications. This ensures that both the depth and semantic maps are consistent and accurate, restoring the physical integrity of the dataset.
A detailed description of the post-processing procedure is provided in the supplement.

\section{Experiments}

\begin{figure}[!t]
	\centering
	\includegraphics[width=0.99\columnwidth]{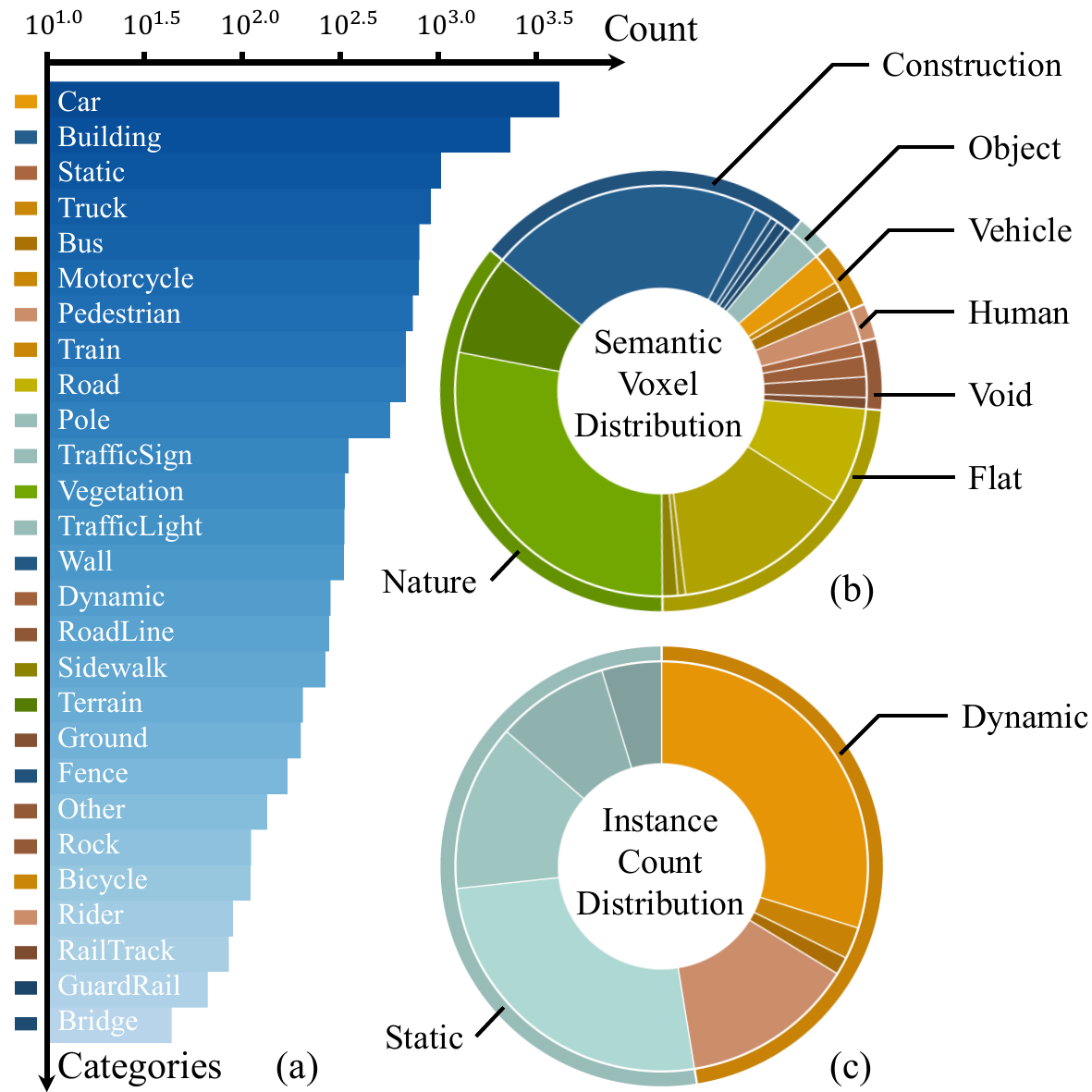}
	\caption{Statistics of ADMesh and CarlaOcc: (a) semantic mesh distribution in ADMesh, (b) semantic voxel distribution in CarlaOcc, and (c) instance count distribution in CarlaOcc. }
	\label{fig.exp_stat}
	%	\vspace{-1.0 em}
\end{figure}

\begin{figure}[!t]
	\centering
	\includegraphics[width=0.99\columnwidth]{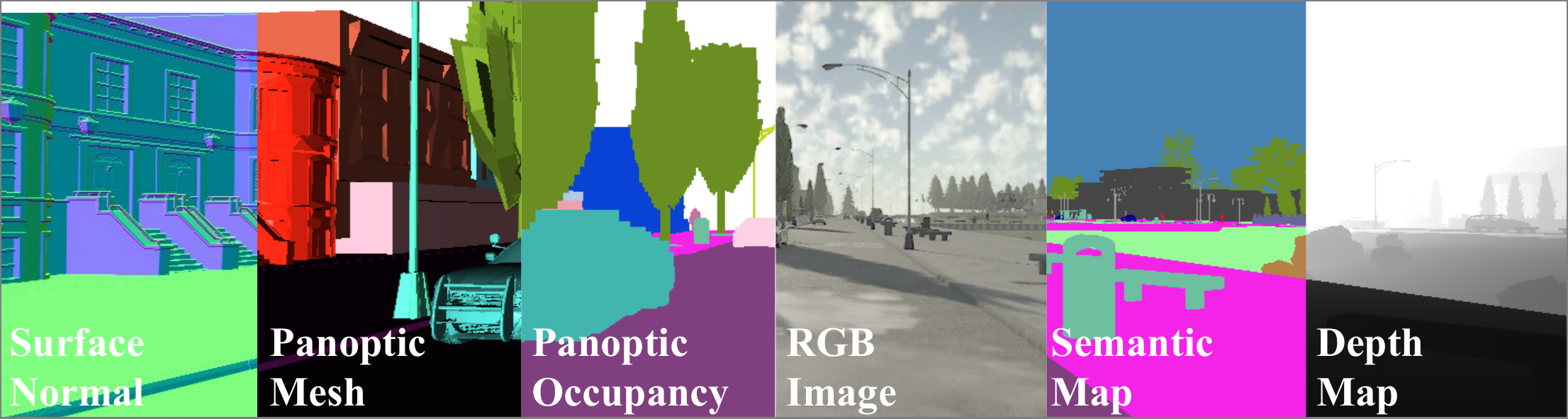}
	\caption{Visualization of data modalities in CarlaOcc.}
	\label{fig.exp_data_modal}
%	\vspace{-1.0 em}
\end{figure}

\begin{figure}[!t]
	\centering
	\includegraphics[width=0.99\columnwidth]{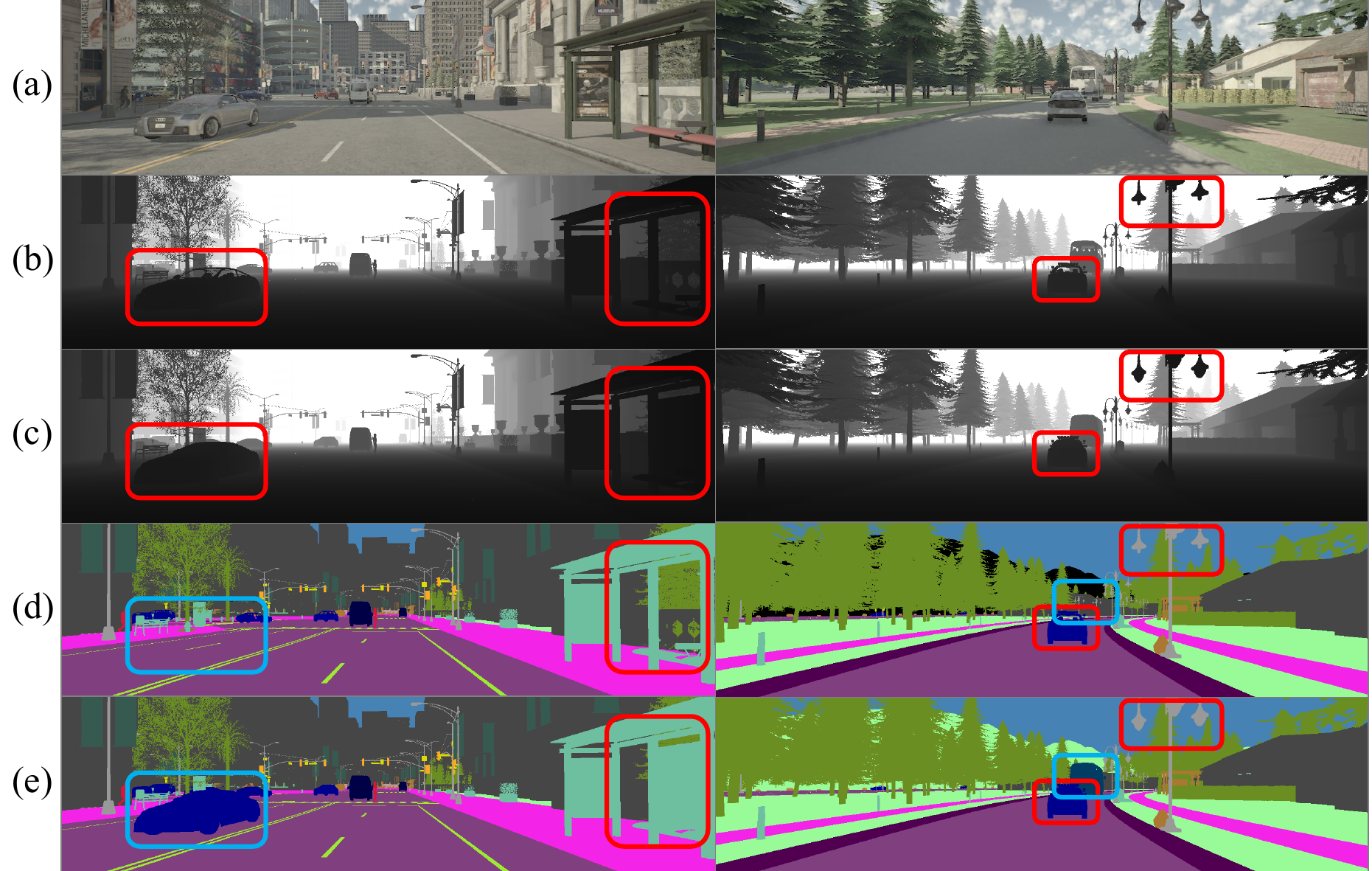}
	\caption{
		Visualizations of instance-guided rectification of sensor artifacts: (a) RGB images; (b) and (d) the rendered depth and semantic maps from CARLA; (c) and (e) the corresponding refined results. Regions with incorrect semantics or depth values caused by transparency are highlighted in red, while the instances that lose their categorical labels are highlighted in blue.}
	\label{fig.sensor_rect}
%	\vspace{-1.0 em}
\end{figure}

\subsection{Dataset Statistics}
Fig.~\ref{fig.exp_stat} provides an overview of the semantic and instance-level characteristics of our benchmark. 
As shown in (a), the semantic meshes in the ADMesh library span a broad set of categories commonly observed in real-world driving scenes. 
(b) depicts the semantic label distribution of the occupancy ground truth in CarlaOcc, where large portions of the 3D space are dominated by construction and nature regions, whereas vehicles, humans, flat areas, and other objects occupy comparatively smaller volumes. 
As illustrated in (c), vehicles and buildings constitute the majority of annotated instances, with static and dynamic categories accounting for 54.2\% and 45.8\% of all instances, respectively. 
These statistics collectively highlight the diverse spatial composition and rich instance variability present in our benchmark, which provides a comprehensive foundation for evaluating semantic and panoptic 3D occupancy prediction tasks.

As shown in Fig.~\ref{fig.exp_data_modal}, CarlaOcc provides physically consistent multi-modal data for camera-centric perception research. Specifically, the RGB images, depth, semantics, LiDAR and semantic LiDAR point cloud are directly collected from CARLA simulator. However, the rendered depth and semantic maps frequently contain artifacts caused by transparency and rendering imperfections. Fig.~\ref{fig.sensor_rect} illustrates our instance-guided rectification process, which corrects erroneous depth and semantic values and restores missing category labels by applying ray casting. The surface normal maps are computed from the rectified depth images using an of-the-shelf depth-to-normal translator D2NT~\cite{feng2023d2nt}. Additionally, we also provides rich instance-level annotations, including positions, velocities, 3D bounding boxes for vehicles and pedestrians, pedestrian gait states, and other dynamic scene attributes. Collectively, these modalities and annotations constitute a rich and unified supervision resource that enables a wide range of tasks in instance-centric 3D perception.

\subsection{Occupancy Dataset Quality Evaluation}
\begin{table}[t]
	\centering
	\settablefont
	\setlength{\tabcolsep}{10pt}
	\caption{Comparison between CarlaOcc and other public occupancy prediction datasets on occupancy quality metrics.}
	\label{tab.dataset_quality}
	\begin{tabular}{l|ccc}
		\toprule[1pt]
		Dataset & Synthetic & $s_{sc}$ & $s_{tc}$ \\
		\midrule
		SemanticKITTI~\cite{semantickitti2019behley}	& $\times$		& 0.353	& 0.023	\\
		KITTI-360-SSCBench~\cite{sscbench2024li} 		& $\times$		& 0.658	& 0.402	\\
		Occ3D-nuScenes~\cite{occ3d2023tian} 			& $\times$		& 0.721	& 0.431	\\
		SurroundOcc~\cite{surroundocc2023wei} 			& $\times$		& 0.878	& 0.589	\\
		CarlaSC~\cite{motionsc2022wilson} 				& $\checkmark$	& 0.887	& 0.775	\\
		\midrule
		\textbf{CarlaOcc (Ours)} 						& $\checkmark$	& \textbf{0.996}	& \textbf{0.873} 	\\
		\bottomrule[1pt]
	\end{tabular}
%	\vspace{-1.0 em}
\end{table}
To evaluate the quality of an occupancy dataset,  we introduce two complementary metrics that quantitatively capture the spatiotemporal coherence of occupancy ground truth.

\noindent\textbf{Spatial Continuity Score.} To evaluate the spatial coherence of occupancy annotations, we define the \emph{Spatial Continuity Score} based on the distribution of isolated occupied voxels at semantic level. 
Intuitively, high-quality occupancy ground truth should exhibit locally continuous spatial structures, where occupied voxels belonging to the same semantic category form connected regions rather than fragmented or isolated points.

Given an occupancy grid $\mathbf{O}_{t}\in\mathbb{Z}^{X\times Y\times Z}$, we first identify the set of occupied voxels $\mathcal{V}_{t}^{(c)}=\{v\mid \mathbf{O}_{t}(v)=c\}$ for each semantic class $c$. The corresponding isolated voxel set $\mathcal{I}_{t}^{(c)}$ is defined as follows:
\begin{equation}
	\mathcal{I}_{t}^{(c)} =
	\{ v \in \mathcal{V}_{t}^{(c)} \mid
	\forall u \in \mathcal{N}(v),\, \mathbf{O}(u) \neq c \},
\end{equation}
where $\mathcal{N}(v)$ denotes the spatially connected 6-neighborhood. The overall Spatial Continuity Score $s_{sc}$ is then defined as the continuity ratio across all classes over the entire dataset:
\begin{equation}
	s_{sc} = 1 -
	\frac{\sum_{t=1}^{T} \sum_{c=1}^{C} |\mathcal{I}_{t}^{(c)}|}
	{\sum_{t=1}^{T} \sum_{c=1}^{C} |\mathcal{V}_{t}^{(c)}|}.
\end{equation}
A higher $s_{sc}$ indicates stronger spatial coherence, implying fewer isolated or fragmented voxels within the semantic occupancy ground truth.

\noindent\textbf{Temporal Consistency Score.}
To assess the temporal stability of occupancy annotations, we define the \emph{Temporal Consistency Score} based on the inter-frame semantic IoU of ground-truth occupancy labels.
Intuitively, consecutive frames within a dynamic scene should exhibit temporally coherent occupancy distributions, where voxel-level semantic states evolve smoothly over time without abrupt changes or inconsistencies.

Given two consecutive semantic occupancy grids and their LiDAR transformations $\mathbf{T}_{t}$ and $\mathbf{T}_{t+1}$, we warp the first grid indices into the second frame and sample the corresponding semantic labels:
\begin{equation}
	\widetilde{\mathcal{V}}_{t}^{(c)}
	=\big\{\Pi\big(\mathbf{T}_{t}\mathbf{T}_{t+1}^{-1}\,x(v), \mathcal{V}_{t+1}^{(c)}\big)\; \big|\; v\in \mathcal{V}_{t}^{(c)}\big\},
\end{equation}
where $x(v)$ denotes the voxel center in metric coordinates and $\Pi$ denotes the grid sampling function.
To avoid penalizing dynamic objects and regions that are newly visible or occluded, the evaluation is restricted to a valid occupancy mask $M_t$.
The overall Temporal Consistency Score is then defined as follows:
\begin{equation}
	s_{tc} =
	\frac{
		\sum_{t=1}^{T-1}\sum_{c=1}^{C} | \widetilde{\mathcal{V}}_{t}^{(c)} \cap \mathcal{V}_{t}^{(c)} \cap M_t |
	}
	{\sum_{t=1}^{T-1}\sum_{c=1}^{C}  |
		\widetilde{\mathcal{V}}_{t}^{(c)} \cup \mathcal{V}_{t}^{(c)} \cap M_t |
	}.
\end{equation}
A higher $s_{tc}$ indicates stronger temporal consistency, reflecting smoother semantic evolution and fewer abrupt occupancy transitions across time.

\noindent\textbf{Dataset Quality Comparison.}
Table~\ref{tab.dataset_quality} summarizes the spatial and temporal coherence of CarlaOcc in comparison with existing public occupancy datasets. Traditional real-world datasets such as SemanticKITTI~\cite{semantickitti2019behley} and KITTI-360-SSCBench~\cite{sscbench2024li} suffer from noticeable fragmentation and temporal discontinuities due to LiDAR sparsity and annotation limitations, leading to low spatial continuity scores and poor temporal consistency.
Synthetic datasets, represented by CarlaSC~\cite{motionsc2022wilson}, exhibit better temporal continuity but still contain non-negligible isolated voxels.
In contrast, CarlaOcc achieves substantially higher coherence across both metrics, improving spatial continuity by about \textbf{12\%} and temporal consistency by around \textbf{13\%} over the existing best dataset. This demonstrates that our dataset provides significantly cleaner, smoother, and more physically consistent occupancy ground truth, benefiting learning-based occupancy prediction models and enabling more reliable evaluation.

\subsection{Occupancy Prediction} 
We benchmark representative semantic and panoptic occupancy prediction methods on the proposed CarlaOcc dataset. 
All experiments adopt a unified voxel size of 0.2\,m.
During evaluation, we report Intersection-over-Union (IoU) and mean Intersection-over-Union (mIoU) in semantic occupancy prediction evaluation~\cite{surroundocc2023wei}, while panoptic occupancy prediction is assessed using Panoptic Quality (PQ), Segmentation Quality (SQ), Recognition Quality (RQ), and mIoU to jointly measure the semantic segmentation quality and instance-level completeness~\cite{panoptic2019kirillov}.

% =============== Panoptic Occupancy Prediction =============== 
\begin{table}[!t]
	\centering
	\settablefont
	\setlength{\tabcolsep}{8pt}
	\caption{Comparison of panoptic occupancy prediction methods on the CarlaOcc dataset.}
	\label{tab.panoptic_occ}
	\begin{tabular}{l|cccc}
		\toprule[1pt]
		Method & PQ $\uparrow$ & SQ $\uparrow$ & RQ $\uparrow$ & mIoU $\uparrow$ \\
		\midrule
		SparseOcc~\cite{sparseocc2024liu} 		& 10.3 & 48.8 & 21.1  & 16.5 \\
		Panoptic-FlashOcc~\cite{flashocc2024yu} 	& 13.5 & 49.1 & 27.5  & 18.4 \\
		\bottomrule[1pt]
	\end{tabular}
	\vspace{-0.5 em}
\end{table}

% =============== Semantic Occupancy Prediction =============== 
\begin{table}[!t]
	\centering
	\setlength{\tabcolsep}{10pt}
	\settablefont
	\caption{Comparison of semantic occupancy prediction methods on the CarlaOcc dataset.}
	\label{tab.semantic_occ}
	\begin{tabular}{c|l|cc}
		\toprule[1pt]
		View & Method & IoU $\uparrow$ & mIoU $\uparrow$ \\
		\midrule
		\multirow{3}{*}{Surround} 
		& GaussianFormer~\cite{gaussianformer2024huang}   & 35.8 & 19.2 \\
		& GaussianFormer2~\cite{gaussianformer2}          & 38.3 & 20.7 \\
		& OPUS~\cite{opus2024wang}                        & 37.0 & 19.3 \\
		\midrule
		\multirow{5}{*}{Single}
		& VoxFormer~\cite{voxformer2023Li}				  & 28.5 & 13.9	 \\
		& OccFormer~\cite{occformer2023Zhang}			  & 30.1 & 14.4	 \\
		& COTR~\cite{cotr2024ma}                          & 31.1 & 15.4	 \\
		& Symphonies~\cite{symphonize2024Jiang}           & 32.6 & 15.9 \\
		& CGFormer~\cite{cgformer2024Yu}                  & 35.2 & 17.8 \\
		\bottomrule[1pt]
	\end{tabular}
	\vspace{-1 em}
\end{table}

\noindent\textbf{Panoptic Occupancy Prediction.}
Due to the early stage of research on this task, only a few existing approaches are capable of performing panoptic occupancy prediction. 
As shown in Table~\ref{tab.panoptic_occ}, all methods exhibit relatively low PQ scores, reflecting the intrinsic difficulty of jointly predicting voxel-wise semantics and instance identities in complex 3D environments. 
Panoptic-FlashOcc~\cite{flashocc2024yu} achieves the highest PQ, outperforming the previous best method PanoOcc~\cite{panoocc2024wang} by 1.6\% and 1.1\% points on PQ and mIoU, respectively. 
SparseOcc~\cite{sparseocc2024liu} demonstrates weaker performance due to limited instance discrimination, especially for small or heavily occluded objects. 
These results highlight the substantial challenge of panoptic occupancy prediction and underscore the need for future research on instance-centric 3D reasoning supported by CarlaOcc.

\noindent\textbf{Semantic Occupancy Prediction.} 
As shown in Table~\ref{tab.semantic_occ}, surround-view methods generally outperform their single-view counterparts in mIoU, highlighting the advantage of multi-view geometric consistency for dense 3D reasoning. 
Meanwhile, single-view models achieve competitive IoU, suggesting that modern architectures can still extract meaningful geometric cues even from limited observations. 

Notably, the overall performance of existing methods on CarlaOcc is slightly lower than their reported results on prior occupancy benchmarks such as SSCBench and SemanticKITTI. 
The performance drop mainly stems from the high-fidelity ground truth in CarlaOcc, which includes precise geometry and faithful occlusions, substantially increasing the learning difficulty.
It further highlights the challenge of our benchmark: models should not only capture the geometric structure of visible regions but also infer the complete 3D layout in occluded areas.

%\noindent\textbf{Domain Adaptation.} 

\section{Conclusion}
We presented ADMesh, a unified 3D mesh library tailored for autonomous driving, and CarlaOcc, a large-scale, physically consistent panoptic occupancy dataset built upon it. Together with standardized evaluation metrics and comprehensive benchmarking, our work establishes a scalable and instance-centric foundation for advancing semantic and panoptic 3D occupancy prediction. We hope this benchmark will facilitate fair comparison, reproducible research, and future progress in holistic 3D scene understanding.

\clearpage
\appendix
\section{Additional Information of CarlaOcc}

\subsection{Data Collection}
To construct a comprehensive dataset for evaluating autonomous driving perception models, we collected diverse and realistic driving sequences distributed across eight CARLA towns.
Each sequence is recorded under a unique combination of town map and traffic configuration, ensuring wide coverage of road topologies, urban structures, and dynamic agent interactions.

\noindent\textbf{Traffic Level Configuration.} 
To simulate realistic and controllable driving environments, we defined four standardized traffic levels: no traffic, light traffic, medium traffic, and heavy traffic. 
The no-traffic setting contains no surrounding vehicles or pedestrians and is included specifically to support methods that require purely static scenes where only the ego vehicle is in motion.
The remaining three levels introduce progressively denser and more complex driving environments by increasing the number of spawned vehicles and pedestrians while tightening their inter-vehicle spacing and increasing speed variability.
These traffic configurations collectively shape the density, aggressiveness, and interaction frequency of dynamic agents, producing scenarios that range from free-driving conditions to highly congested urban environments.

\noindent\textbf{Sequence Distribution.} 
The collected 104 sequences are evenly distributed across 8 CARLA towns, with each town containing one sequence for no-traffic configuration, three sequences for low and medium traffic configuration, and five sequences for high traffic configuration. 
To ensure balanced spatial diversity, we use Farthest Point Sampling to initialize each sequence with a distinct starting position and driving route. This strategy remarkably prevents redundant scene coverage across sequences and maximizes the utilization of spatial diversity in each town.

\noindent\textbf{Sensor Configuration.}
Our sensor setup follows the KITTI-360~\cite{kitti360_2022liao} configuration for cameras 00-03 and the LiDAR, including their viewing directions, intrinsic parameters, relative transformations, and the LiDAR mounting height. However, the original KITTI-360 design presents several limitations for occupancy prediction. First, cameras 00 and 01 are tilted downward by $5^\circ$, causing the camera-frame occupancy to intersect the ground prematurely and disrupting the geometric consistency required for single-view occupancy learning. Second, these two cameras use fisheye lenses and require additional rectification to match the perspective cameras, which complicates the training process. Third, KITTI-360 lacks a rear-view camera, making full surround-view occupancy prediction infeasible. 

\begin{figure*}[!t]
	\centering
	\includegraphics[width=0.75\textwidth]{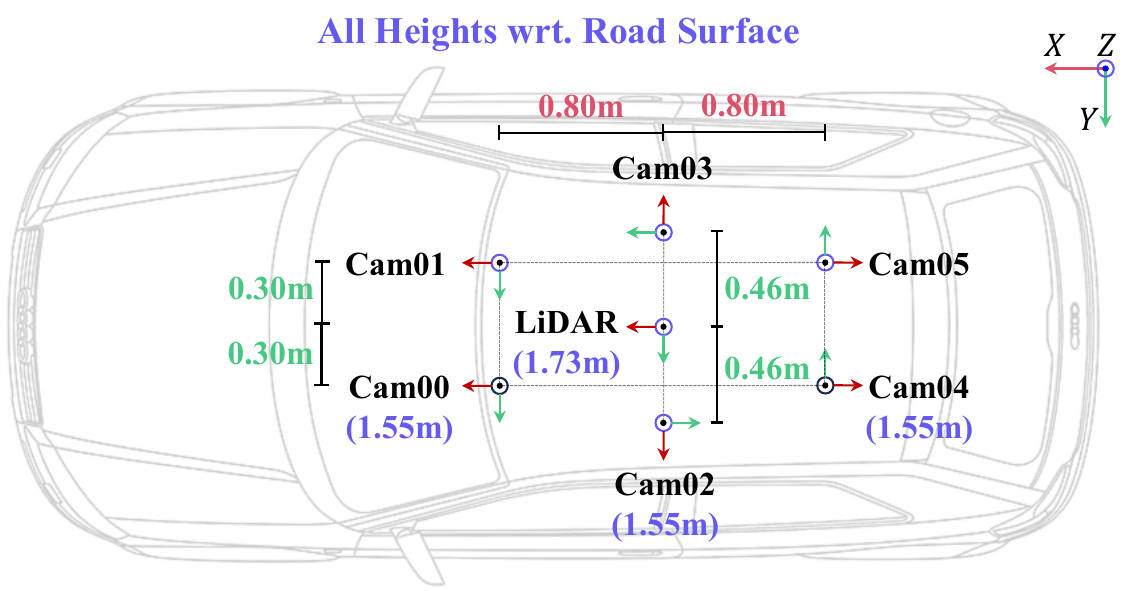}
	\caption{Top-down illustration of the sensor configuration in the CarlaOcc dataset.}
	\label{fig.sensor_cfg}
\end{figure*}

To address these issues, we redesign the sensor suite by removing the camera tilt, using perspective cameras for all views, and adding rear-view cameras to achieve a complete 360$^\circ$ coverage. As illustrated in Fig.~\ref{fig.sensor_cfg}, cameras 00-05 form a symmetric surround-view setup, each providing RGB, semantic segmentation, and depth maps with a unified field of view of 103.7$^\circ$ and the same resolution of 1,408$\times$376 pixels. The 32-channel LiDAR operates with a maximum range of 80\,m and a rotation frequency of 20\,Hz, producing approximately 250,000 points per second. Its vertical field of view spans from -30$^\circ$ to 10$^\circ$, providing dense coverage of both near-ground geometry and elevated structures.

\subsection{Skeletal Motion Analyzer}
To enable precise reconstruction of deforming pedestrian meshes, we develop a \textit{Skeletal Motion Analyzer} that aligns real-time skeletal motion observations with a pre-recorded canonical walking cycle.

\noindent\textbf{Canonical Gait Database.}
We construct a standard gait database $\mathcal{G} = \{\mathbf{g}_d\}_{d=0}^{D-1}$ containing $D$ discrete phases of a complete walking cycle. 
Each entry $\mathbf{g}_d$ stores the transformation matrices for key articulations and the extracted kinematic descriptors.
For each phase index $d$, we compute a motion characteristic descriptor
\begin{equation}
	\delta_d = 
	\begin{bmatrix}
		\Delta x^d \\[2pt]
		\theta_{\mathrm{rel}}^d
	\end{bmatrix},
\end{equation}
where $\Delta x^d$ encodes the forward displacement differential and $\theta_{\mathrm{rel}}^d$ denotes the relative articulation angle derived from joint rotations.

\noindent\textbf{Real-time Motion Analysis.}
For each pedestrian instance $k$ at frame $t$, we extract a local temporal window of $W$ frames:
\begin{equation}
	\mathcal{X}_k^t = \left\{ \mathbf{x}_k^{t-\frac{W}{2}}, \dots, \mathbf{x}_k^{t}, \dots, \mathbf{x}_k^{t+\frac{W}{2}} \right\},
\end{equation}
where each observation $\mathbf{x}_k^\tau$ contains instantaneous kinematic features. To align the observed motion with canonical gait phases, we derive an observation descriptor for the central frame:
\begin{equation}
	\hat{\delta}_k^t =
	\begin{bmatrix}
		\Delta \hat{x}_k^t \\[2pt]
		\hat{\theta}_{\mathrm{rel},k}^t
	\end{bmatrix},
\end{equation}
where $\Delta \hat{x}_k^t$ is estimated from the temporal window and $\hat{\theta}_{\mathrm{rel},k}^t$ is obtained from articulated joint rotations.

We then compute a phase likelihood score for each canonical phase $d$ as
\begin{equation}
	\mathcal{L}(d \mid \hat{\delta}_k^t)
	= \exp\!\left( - \frac{1}{2}
	\left\| \hat{\delta}_k^t - \delta_d \right\|_{\Sigma^{-1}}^2 \right),
\end{equation}
where $\Sigma$ is a diagonal covariance matrix calibrated from training data.  
The matched gait phase is obtained via maximum-likelihood estimation:
\begin{equation}
	d_k = \arg\max_{d \in [0, D-1]} \mathcal{L}(d \mid \hat{\delta}_k^t).
\end{equation}

\noindent\textbf{Mesh Retrieval and Reconstruction.}
Once the phase $d_k$ is determined, the corresponding posed mesh $\mathcal{M}_{d_k}$ is retrieved from the canonical gait database and transformed to the world coordinate system using the recorded global transformation $\mathbf{T}_k$, yielding a geometrically accurate reconstruction of the pedestrian's instantaneous pose. This phase matching approach achieves robust gait alignment without requiring explicit skeletal retargeting or expensive inverse kinematics, enabling efficient processing of large-scale pedestrian populations in simulation.

\subsection{Sensor Artifacts Rectification}
As discussed in Sec.~3.2 in the main paper, the multi-modal signals rendered in Unreal Engine~5 frequently exhibit systematic artifacts that violate geometric and semantic consistency. These issues mainly arise from the way that rendering pipeline handles transparent (\textit{e.g.}, glass, windows) and non-Nanite\footnote{Nanite is Unreal Engine~5's virtualized micropolygon geometry system that enables efficient rendering of extremely high-detail meshes.} (\textit{e.g.}, terrain) materials:
\begin{itemize}
	\item Translucent materials are excluded from the depth buffer, which causes the rendered depth to correspond to the first opaque surface behind the transparent object.
	\item The semantic rendering pass, inherited from the legacy CARLA implementation, applies material substitution only to the Nanite rendering path.
\end{itemize}
As a result, transparent objects inherit the depth and semantics of background structures, and some non-Nanite instances disappear entirely from the semantic maps.
These modality-specific failures manifest as broken silhouette alignment, incorrect depth ordering, and incomplete semantic regions, thereby corrupting the physical integrity of the dataset.

To address these issues, we introduce an instance-guided rectification method that reconstructs per-frame, physically consistent depth and semantic maps by using instance-level geometry priors, without the need of modifying the rendering engine. 

\noindent\textbf{Depth Map Rectification.}
We reconstruct per-camera, ray-casted geometry and fuse it with the sensor-collected depths in a physically consistent manner.
For each frame, we first assemble a compact scene mesh $\mathcal{M}$ containing only the geometries of transparent or semi-transparent objects, as well as those exhibiting semantic inconsistencies in the rendered maps. The remainder of the static environment is excluded to mitigate ray-casting overhead and to prevent artifacts caused by texture-dependent meshes (\textit{e.g.}, vegetation), whose system-exported geometry is often less detailed than the sensor-collected signals.
Subsequently, 
Then, by using the calibrated camera intrinsics and extrinsics, we create a tensor-accelerated pinhole ray bundle, where each ray originates from the optical center $\mathbf{o}_{\text{cam}}$ and travels along the viewing direction $\mathbf{d}(u,v)$. The ray-casted depth can be obtained from the distance to the first visible surface intersected by each ray:
\begin{equation}
	\mathbf{D}_{\text{rc}}(u,v) = \min_{t>0}\{\,t \mid \mathbf{o}_{\text{cam}} + t\,\mathbf{d}(u,v) \in \mathcal{M}\,\}.
\end{equation}
Subsequently, the raw depth map $\mathbf{D}_{\text{raw}}$ is rectified by taking the pointwise minimum with the ray-casted result:
\begin{equation}
	\mathbf{\hat{D}}(u,v) = \min\big(\mathbf{D}_{\text{raw}}(u,v),\, \mathbf{D}_{\text{rc}}(u,v)\big),
\end{equation}
which effectively restores the correct surface geometry for transparent and semi-transparent regions that were incorrectly assigned to the background, while preserving valid measurements elsewhere. This depth fusion strategy ensures that each pixel records the nearest physically consistent surface along its viewing ray, thereby improving geometric fidelity and cross-modal consistency.

\begin{figure*}[!t]
	\centering
	\includegraphics[width=0.99\textwidth]{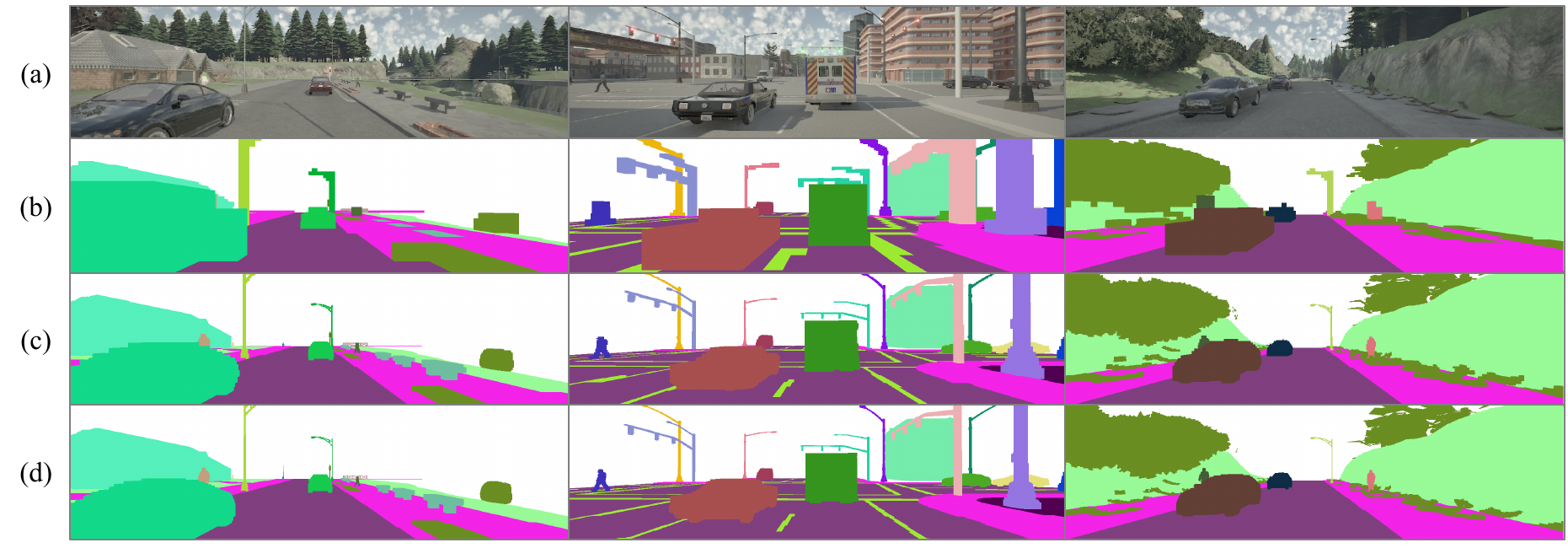}
	\caption{Visualization of occupancy ground truth at different voxel resolutions: (a) RGB images, (b) panoptic occupancy labels at 0.5\,m voxel size, (c) 0.1\,m voxel size, and (d) 0.05\,m voxel size. }
	\label{fig.vis_multires}
\end{figure*}

\noindent\textbf{Semantic Map Rectification.}
Unlike depth rectification which enforces geometric consistency through the nearest-surface fusion, semantic rectification focuses on restoring categorical correctness and instance completeness. Given the mesh subset $\mathcal{M}$ described above, we first generate a ray-casted semantic map $\mathbf{S}_{\text{rc}}$ by querying, for each camera ray, the semantic label of the first intersected triangle within $\mathcal{M}$. This produces physically grounded labels for all transparent or inconsistent regions, where the collected semantic map $\mathbf{S}_{\text{raw}}$ often misattributes the semantics of the background objects.

To identify the degraded regions, we exploit the depth consistency between the previously rectified depth $\mathbf{\hat{D}}$, the collected depth $\mathbf{D}_{\text{raw}}$, and the ray-casted depth $\mathbf{D}_{\text{rc}}$. 
Pixels with significant depth discrepancies indicate that the collected semantics are associated with surfaces behind transparent materials rather than the true foreground geometry. Therefore, we define the transparency region $\Omega_{\text{t}}$ as follows:
\begin{equation}
	\Omega_{\text{t}}(u,v) = \mathbbm{1}\!\left[\mathbf{D}_{\text{raw}}(u,v) - \mathbf{\hat{D}}(u,v) \ge \varepsilon\right],
\end{equation}
where $\mathbbm{1}\left[\cdot\right]$ denotes the indicator function, and $\varepsilon$ is a small positive threshold that distinguishes meaningful depth differences from sensor noise. This condition captures pixels where the collected depth erroneously penetrates through transparent surfaces to record background geometry.
In addition to transparency-induced mislabels, certain instances disappear completely in $\mathbf{S}_{\text{raw}}$ due to missing or invalid semantic substitution during rendering. We detect such omission region $\Omega_{\text{s}}$ by finding unlabeled pixels in the collected map:
\begin{equation}
	\Omega_{\text{s}}(u,v) = \mathbbm{1}\!\left[\mathbf{S}_{\text{raw}}(u,v) = s_\varnothing\right],
\end{equation}
where $s_\varnothing$ denotes the empty label assigned to unrendered instances. These regions typically correspond to non-Nanite assets that are excluded from the semantic rendering pipeline, creating semantic voids in otherwise geometrically complete areas.
The complete error region $\Omega_{\text{err}} = \Omega_{\text{t}} \cup \Omega_{\text{s}}$ for semantic rectification is then defined as the union of transparency-induced mislabels and omission artifacts, and the final rectified semantic map $\mathbf{\hat{S}}$ is obtained by strategically replacing semantics in these error regions:
\begin{equation}
	\mathbf{\hat{S}}(u,v)=
	\begin{cases}
		\mathbf{S}_{\text{rc}}(u,v), & (u,v)\in\Omega_{\text{err}}, \\
		\mathbf{S}_{\text{raw}}(u,v), & \text{otherwise}.
	\end{cases}
\end{equation}

This unified rectification framework simultaneously addresses both transparency-induced semantic misattribution and instance omission artifacts, ensuring that all visible surfaces receive physically consistent and categorically accurate depth measurements and semantic labels. The approach effectively bridges the gap between the detailed geometry captured by the depth sensor and the categorical information provided by the semantic renderer, producing physically consistent depth and semantic outputs suitable for training robust perception systems.

\begin{figure*}[!t]
	\centering
	\includegraphics[width=0.99\textwidth]{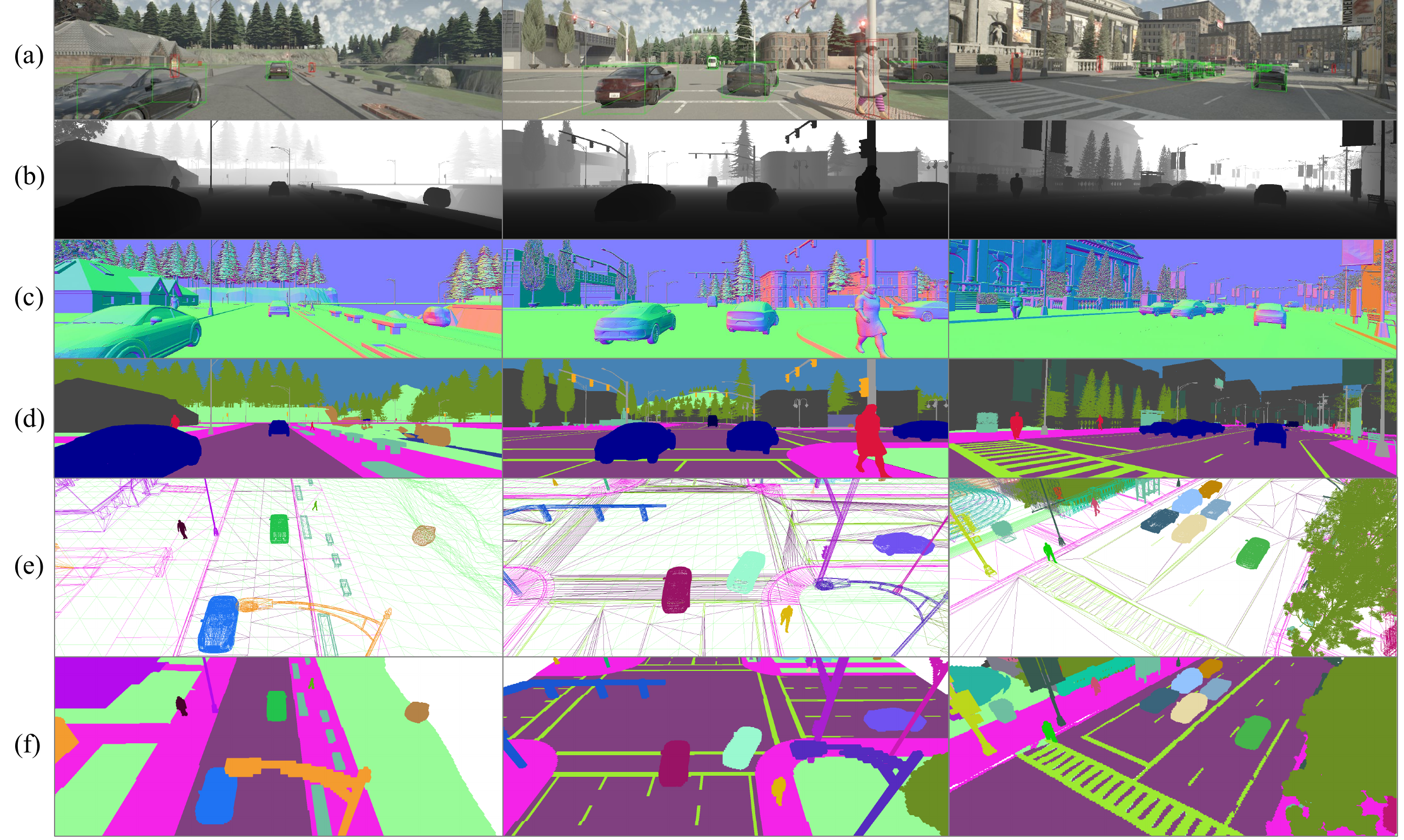}
	\caption{Visualization of multi-modal data in CarlaOcc: (a) RGB images with rich annotations, (b) depth maps, (c) surface normal maps, (d) semantic segmentation maps, (e) panoptic meshes shown in wireframe, and (f) panoptic occupancy ground truth. ``Stuff'' classes are visualized using the official color map, while ``thing'' instances are displayed in randomly assigned colors.}
	\label{fig.data_vis}
\end{figure*}

\begin{figure*}[!t]
	\centering
	\includegraphics[width=0.99\textwidth]{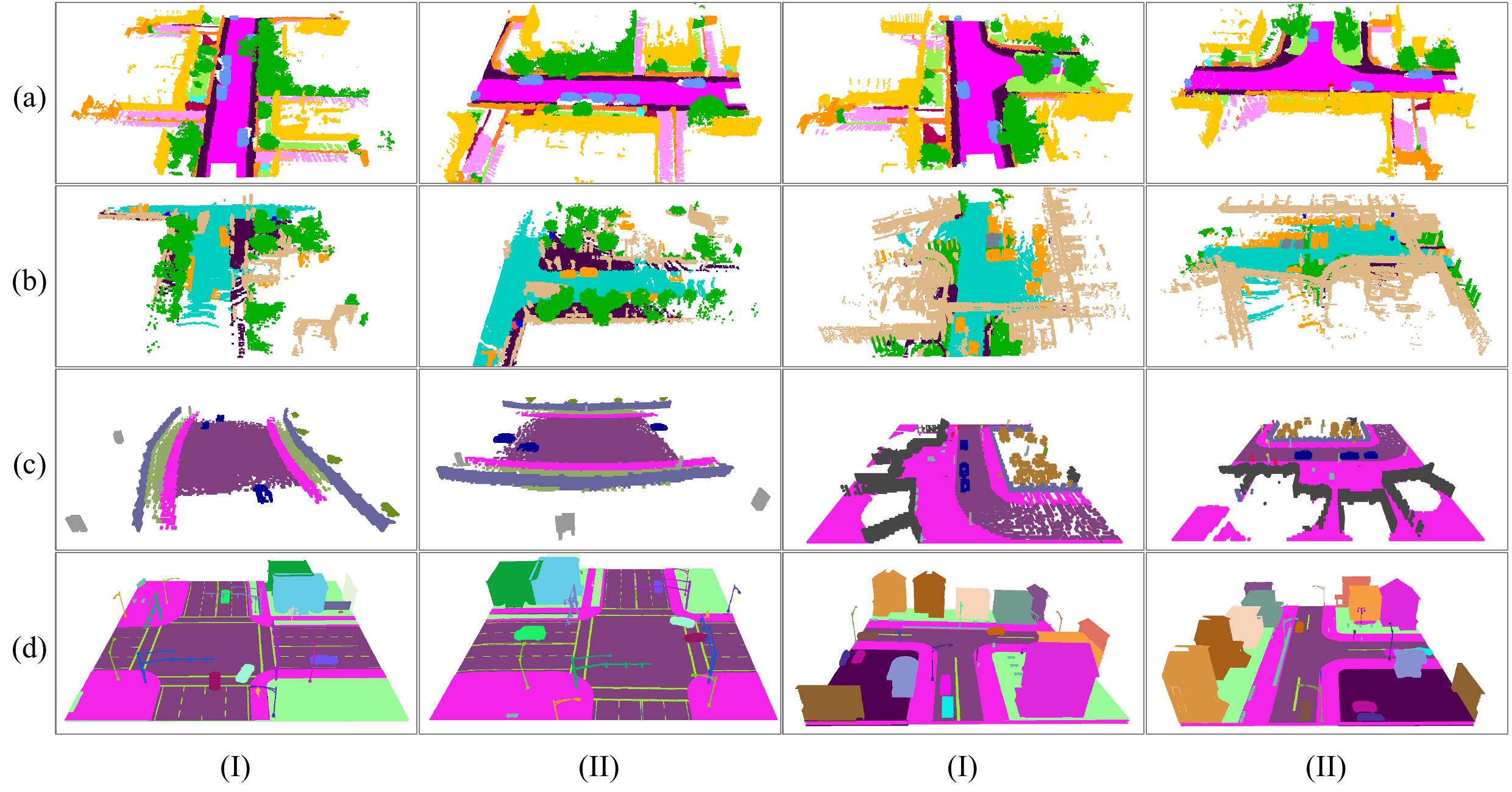}
	\caption{Ground truth quality comparison between existing public occupancy datasets and CarlaOcc: (a) KITTI-360-SSCBench~\cite{sscbench2024li}, (b) Occ3D-nuScenes~\cite{occ3d2023tian}, (c) CarlaSC~\cite{motionsc2022wilson}, and (d) our proposed CarlaOcc dataset. For each dataset, we present the (I) forward view and the (II) left-side view for better comparison. }
	\label{fig.dataset_comparison}
\end{figure*}

\subsection{Data Visualization.} 
To provide an intuitive understanding of the richness and cross-modal consistency of CarlaOcc, we visualize various samples of the dataset. The dataset offers a diverse set of complementary modalities rendered from the same calibrated multi-camera rig, enabling holistic scene perception. As shown in Fig.~\ref{fig.data_vis}, each frame includes high-fidelity RGB images, metrically accurate depth maps, dense surface normals, semantic segmentation maps, instance-level panoptic meshes, and the resulting panoptic occupancy ground truth. These visualizations highlight the geometric completeness of our reconstructed scenes and the fine-grained, instance-level annotations preserved across modalities.

In Fig.~\ref{fig.vis_multires}, we further demonstrate the panoptic occupancy ground truth generated at different voxel resolutions, ranging from coarse (0.5\,m) to fine-grained (0.05\,m) grids. While coarser voxels provide a compact global overview of the 3D scene layout, reducing memory and computation costs, higher-resolution grids accurately capture object silhouettes, scene boundaries, and small structures such as pedestrians or traffic signs. This multi-resolution supervision enables researchers to adapt model capacity and training objectives to different computational budgets and application scenarios. Together, these visualizations illustrate the high geometric fidelity and scalable annotation quality that CarlaOcc offers for 3D occupancy perception research.

\section{Additional Experimental Results}

\subsection{Occupancy Dataset Quality Comparison}
We compare the quality of occupancy ground truth produced by existing datasets and our proposed CarlaOcc dataset. Traditional pipelines such as KITTI-360-SSCBench~\cite{sscbench2024li} and Occ3D-nuScenes~\cite{occ3d2023tian} construct occupancy labels by directly voxelizing sparse LiDAR point clouds. Since LiDAR measurements are inherently incomplete and noisy, this strategy inevitably leads to broken surfaces, isolated floating points, and large holes in unobserved or distant regions.
As shown in Fig.~\ref{fig.dataset_comparison}(a) and (b), buildings, sidewalks, and road boundaries are clearly fragmented, and many regions that should be occupied are missing, which results in suboptimal and misleading supervisory signals for learning 3D environment layout.

Synthetic datasets like CarlaSC~\cite{motionsc2022wilson} adopt a similar paradigm even though they are built in simulation. They still rely on multiple LiDAR sensors to scan the environment and then discretize the resulting point clouds into voxels. Consequently, the final occupancy labels are still limited by LiDAR sampling density, occlusions, and sensor noise, rather than reflecting the complete underlying scene geometry. This can be observed in Fig.~\ref{fig.dataset_comparison}(c), where road surfaces and curbs remain discontinuous and the geometry of surrounding structures is coarse and irregular.

In contrast, our CarlaOcc dataset bypasses LiDAR sampling and directly accesses the high-fidelity 3D mesh assets within the simulator. We retrieve the full polygonal meshes of all relevant scene elements, and then perform voxelization and precise cropping in the world coordinate system to obtain dense and consistent occupancy labels. As illustrated in Fig.~\ref{fig.dataset_comparison}(d), this mesh-based pipeline produces smooth surfaces, clean object boundaries, and almost no noisy or missing regions in both the forward and left-side views. Overall, CarlaOcc delivers significantly higher-quality occupancy ground truth than prior LiDAR-based pipelines, providing a much stronger supervisory signal for training 3D occupancy prediction models.

\begin{figure}[!t]
	\centering
	\includegraphics[width=0.99\columnwidth]{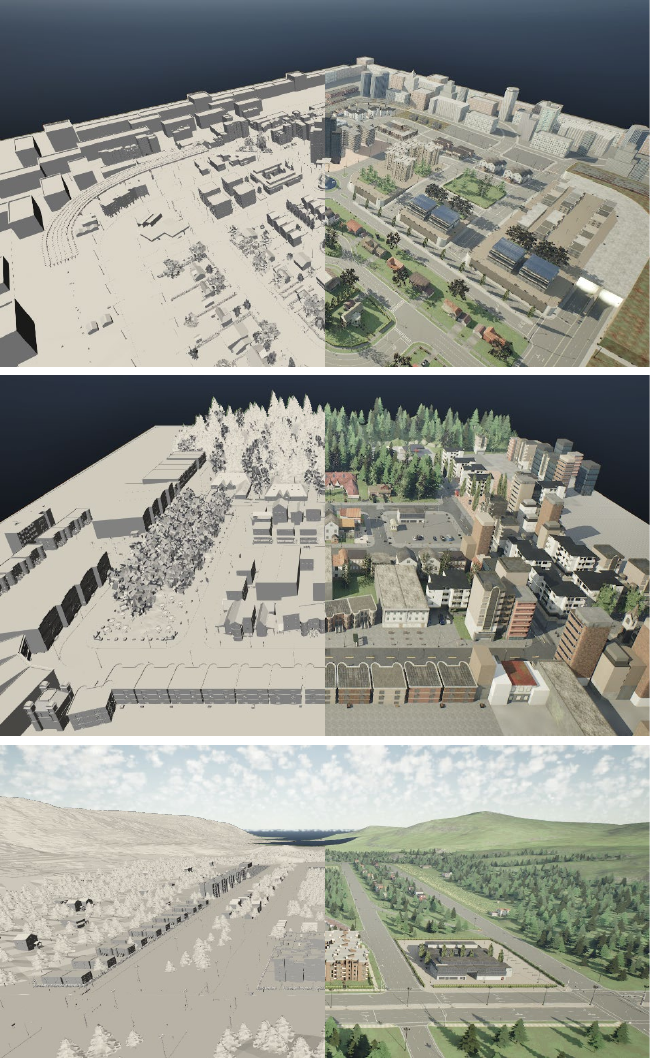}
	\caption{Visualization of the reconstructed CARLA maps. Left: unified static scene meshes obtained through our reconstruction pipeline. Right: corresponding RGB images captured in Unreal Engine 5. }
	\label{fig.town_recon}
\end{figure}

\begin{figure*}[!t]
	\centering
	\includegraphics[width=0.99\textwidth]{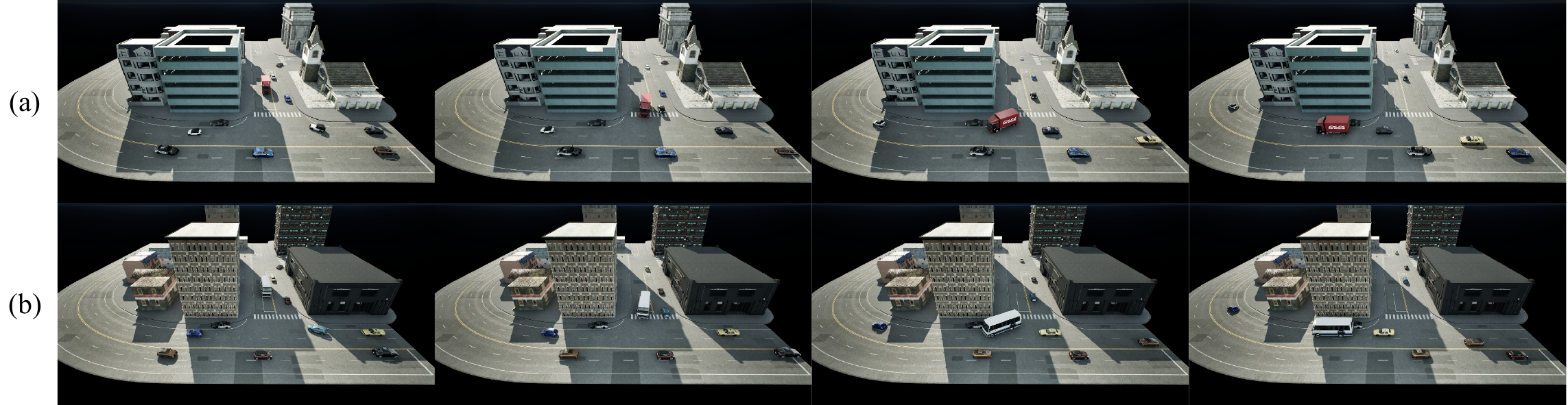}
	\caption{
		Parametric scene generation and editing on ADMesh. Each row shows a dynamic sequence rendered at four successive time steps.
		(a) Original scene generated by assigning trajectories and poses to the	static mesh actors.
		(b) Edited scene obtained by replacing the mesh asset of each actor while keeping all trajectories, camera parameters, and illumination fixed.
		The example illustrates that our parametric representation can generate	diverse, high-quality dynamic urban scenes with simple actor-level edits.
	}
	\label{fig.exp_pcg}
\end{figure*}

\subsection{City-Level Scene Reconstruction}
We merge all exported static meshes of each town into a single unified mesh with their corresponding transformation matrices, yielding a full city-level reconstruction of the CARLA maps. As shown in Fig.~\ref{fig.town_recon}, the resulting reconstructed towns accurately preserve roads, buildings, terrain, and large‑scale layout. The left column visualizes the unified static mesh produced by our pipeline, while the right column shows the corresponding RGB rendering in Unreal Engine 5, confirming that our reconstruction precisely matches the original simulator scenes.

\subsection{Parametric Scene Generation}
We further demonstrate that the proposed ADMesh representation naturally supports
parametric generation of high-quality dynamic scenes. 
Starting from the static mesh library in ADMesh (buildings, roads, vehicles, \textit{etc}.), we describe each scene by a compact set of actor-level parameters:
(i) the index of the mesh asset,
(ii) a rigid transform in the world coordinate system, and
(iii) an optional time-dependent trajectory function that specifies how the actor
moves over time. 
Given a set of trajectories and initial placements that respect the road layout, we can instantiate all actors at each simulation step $t_k$, assign the corresponding transform to their meshes, and render the resulting frame from a fixed virtual camera. By advancing $t_k$ we obtain a temporally coherent, physically plausible dynamic sequence, while the underlying geometry always comes from high-quality meshes in ADMesh.

To evaluate the flexibility of this parametric representation, we construct two types of sequences. 
As shown in Fig.~\ref{fig.exp_pcg}, we fix a configuration of buildings and vehicles and simulate their trajectories over several time steps, yielding a dynamic urban scene in which
cars move along the road while the camera remains static. Second, we keep all trajectories and camera parameters unchanged, but replace the mesh asset assigned to each actor (\textit{e.g.}, replacing building or vehicle models for another with a compatible footprint). 
This simple mesh substitution produces a new dynamic scene with different appearance and layout, while preserving temporal consistency and avoiding any
manual re-authoring of trajectories. The experiment indicates that ADMesh enables efficient scene editing and instance replacement at the level of individual actors, making it suitable for large-scale generation of diverse driving scenarios.

\subsection{3D Occupancy Prediction}
\noindent\textbf{Ablation on Voxel Size.} 
To analyze the impact of voxel resolution on 3D semantic occupancy prediction, we conduct an ablation study on the CarlaOcc dataset by varying the voxel size from 0.05\,m and 0.10\,m to 0.50\,m for Symphonies~\cite{symphonize2024Jiang} and GaussianFormer2~\cite{gaussianformer2}. As shown in Table~\ref{tab.voxel_ablation}, the coarsest voxel size (0.50\,m) achieves the best IoU and mIoU for both methods, while the performance decreases when using finer voxels. This is mainly because CarlaOcc provides high-resolution ground-truth occupancy, where even small localization errors lead to noticeable misalignment with the dense labels at finer voxel resolutions, making IoU and mIoU much more sensitive and the task more challenging.

\noindent\textbf{Per-Class Results on CarlaOcc.} 
In the main paper, we report only the overall IoU and mIoU on the CarlaOcc benchmark. For completeness, Table~\ref{tab.ssc} further provides the per-class IoU for all semantic categories. The distribution of scores is consistent with the overall metrics: road, building and vegetation related classes are generally easier, while small and thin objects such as poles, traffic signs and traffic lights remain more challenging. These results offer a detailed per-class benchmark for CarlaOcc and can serve as a reference for future monocular 3D semantic occupancy methods evaluated on this dataset.

\begin{table}[t]
	\centering
	\settablefont
	\setlength{\tabcolsep}{6pt}
	\caption{Ablation study on voxel size for semantic occupancy prediction on the CarlaOcc dataset.}
	\label{tab.voxel_ablation}
	\begin{tabular}{l|l|cc}
		\toprule[1pt]
		Method & Voxel size (m) & IoU & mIoU \\
		\midrule
		\multirow{3}{*}{Symphonies~\cite{symphonize2024Jiang}} 
		& 0.05 & 15.2 & 10.1 \\
		& 0.10 & 21.5 & 12.1 \\
		& 0.50 & 32.6 & 15.9 \\
		\midrule
		\multirow{3}{*}{GaussianFormer2~\cite{gaussianformer2}} 
		& 0.05 & 18.4 & 10.9 \\
		& 0.10 & 23.1 & 14.8 \\
		& 0.50 & 38.3 & 20.7 \\
		\bottomrule[1pt]
	\end{tabular}
\end{table}

\begin{table}[t]
	\centering
	\small
	\settablefont
	\caption{Sim-to-real evaluation results of representative baselines pretrained on CarlaOcc and finetuned on real-world datasets.}
	\begin{tabular}{lllcc}
		\toprule[1pt]
		\textbf{Pretrain}
		& \textbf{Target}
		& \textbf{Method}
		& \textbf{Scratch mIoU}
		& \textbf{mIoU} \\
		\midrule
		CarlaOcc & KITTI-360 & Symphonies & 15.9 & \textbf{17.4} \\
		CarlaOcc & nuScenes  & SparseOcc  & 16.5 & \textbf{17.3} \\
		\bottomrule[1pt]
	\end{tabular}
	\label{tab:sim2real_results}
\end{table}

\begin{table*}[!t]
	\centering
	\settablefont
	\setlength{\tabcolsep}{4pt}
	\caption{Semantic occupancy prediction results on the CarlaOcc dataset. }
	\label{tab.ssc}
	\begin{tabular}{l|cc|ccccccccccccccc}
		\toprule[1pt]
		Method & IoU & mIoU
		& \rotatebox{90}{\textcolor{roadcolor}{$\blacksquare$}~Road}
		& \rotatebox{90}{\textcolor{sidewalkcolor}{$\blacksquare$}~Sidewalk}
		& \rotatebox{90}{\textcolor{buildingcolor}{$\blacksquare$}~Building}
		& \rotatebox{90}{\textcolor{wallcolor}{$\blacksquare$}~Wall}
		& \rotatebox{90}{\textcolor{fencecolor}{$\blacksquare$}~Fence}
		& \rotatebox{90}{\textcolor{polecolor}{$\blacksquare$}~Pole}
		& \rotatebox{90}{\textcolor{trafficlightcolor}{$\blacksquare$}~TrafficLight}
		& \rotatebox{90}{\textcolor{trafficsigncolor}{$\blacksquare$}~TrafficSign}
		& \rotatebox{90}{\textcolor{vegetationcolor}{$\blacksquare$}~Vegetation}
		& \rotatebox{90}{\textcolor{groundcolor}{$\blacksquare$}~Ground}
		& \rotatebox{90}{\textcolor{personcolor}{$\blacksquare$}~Person}
		& \rotatebox{90}{\textcolor{carcolor}{$\blacksquare$}~Car}
		& \rotatebox{90}{\textcolor{truckcolor}{$\blacksquare$}~Truck}
		& \rotatebox{90}{\textcolor{othervehiclecolor}{$\blacksquare$}~OtherVehicle}
		& \rotatebox{90}{\textcolor{othercolor}{$\blacksquare$}~Other}
		\\	\midrule
		Symphonies~\cite{symphonize2024Jiang} &
		32.6 & 15.9 &
		36.74 & 20.04 & 21.38 & 6.68 & 10.02 & 5.34 &
		6.68 & 6.01 & 26.72 & 16.70 & 5.34 &
		30.06 & 23.38 & 16.70 & 6.71 \\
		SparseOcc~\cite{sparseocc2024liu} &
		30.1 & 14.4 &
		33.28 & 18.15 & 19.36 & 6.05 & 9.08 & 4.84 &
		6.05 & 5.45 & 24.20 & 15.13 & 4.84 &
		27.23 & 21.18 & 15.13 & 6.03 \\
		OPUS~\cite{opus2024wang} &
		37.0 & 19.3 &
		44.60 & 24.33 & 25.95 & 8.11 & 12.16 & 6.49 &
		8.11 & 7.30 & 32.44 & 20.27 & 6.49 &
		36.49 & 28.38 & 20.27 & 8.11 \\
		GaussianFormer2~\cite{gaussianformer2} &
		38.3 & 20.7 &
		47.84 & 26.09 & 27.83 & 8.70 & 13.05 & 6.96 &
		8.70 & 7.83 & 34.79 & 21.74 & 6.96 &
		39.14 & 30.44 & 21.74 & 8.69 \\
		\bottomrule[1pt]
	\end{tabular}
\end{table*}

\begin{table*}[!t]
	\centering
	\settablefont
	\caption{Depth estimation results on the CarlaOcc dataset.}
	\begin{tabular}{l|l|cccc|ccc}
		\toprule[1pt]
		Category & Method & AbsRel $\downarrow$ & SqRel $\downarrow$ & RMSE $\downarrow$ & RMSElog $\downarrow$
		& $\delta <1.25$ $\uparrow$ & $\delta <1.25^2$ $\uparrow$ & $\delta <1.25^3$ $\uparrow$ \\
		\midrule
		\multirow{4}{*}{Discriminative}
		& MiDaS~\cite{midas2023birkl}                 
		& 0.128 & 1.642 & 7.215 & 0.335 & 0.842 & 0.948 & 0.977 \\		
		& LeReS~\cite{leres2021yin}                 
		& 0.112 & 1.392 & 6.932 & 0.314 & 0.868 & 0.957 & 0.980 \\		
		& DPT~\cite{dpt2021ranftl}                   
		& 0.101 & 1.248 & 6.621 & 0.302 & 0.887 & 0.964 & 0.982 \\		
		& DepthAnythingV2~\cite{depthanythingv2_2024yang}       
		& 0.093 & 1.035 & 6.159 & 0.297 & 0.899 & 0.965 & 0.983 \\
		\midrule		
		\multirow{3}{*}{Generative}
		& Marigold~\cite{marigold2024ke}              
		& 0.089 & 1.012 & 6.045 & 0.284 & 0.904 & 0.968 & 0.985 \\		
		& LOTUS~\cite{lotus2024he}                 
		& 0.085 & 0.985 & 5.918 & 0.276 & 0.911 & 0.972 & 0.987 \\
		\midrule		
		\multirow{2}{*}{Metric}
		& Metric3Dv2~\cite{metric3dv2_2024hu}            
		& 0.087 & 1.111 & 6.427 & 0.199 & 0.899 & 0.959 & 0.975 \\
		& UniDepthV2~\cite{unidepthv2_2025piccinelli}            
		& 0.082 & 0.942 & 5.731 & 0.219 & 0.928 & 0.975 & 0.989 \\
		\bottomrule[1pt]
	\end{tabular}
	\label{tab.depth}
\end{table*}

\begin{figure}[t]
	\centering
	\includegraphics[width=0.99\linewidth]{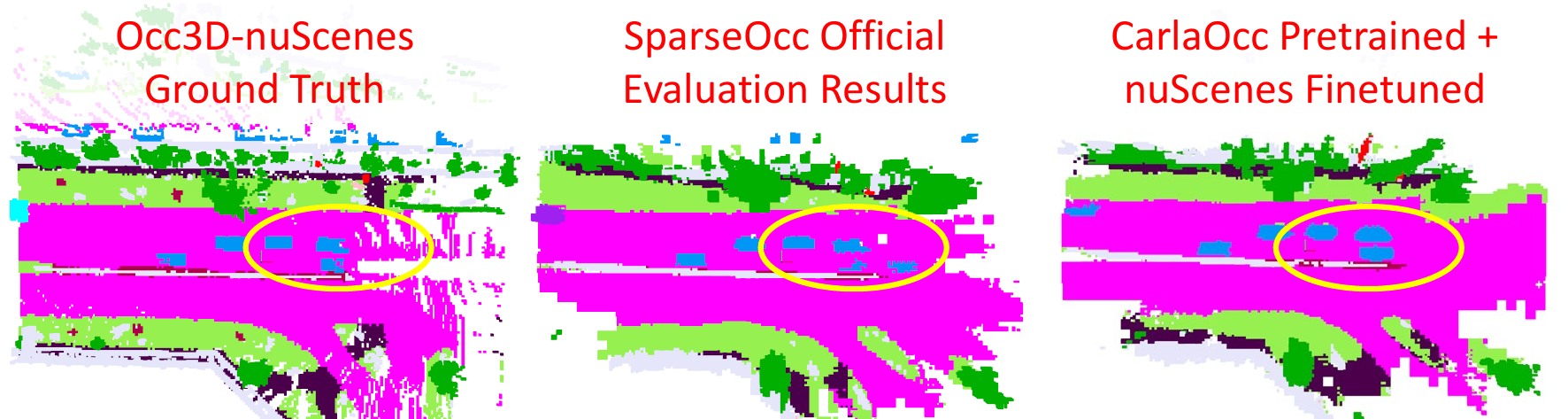}
	\caption{Visualizations of sim-to-real experiment. As shown in yellow circles, SparseOcc exhibits more complete instance contours after pretraining on CarlaOcc compared to official results.}
	\label{fig:comparison}
\end{figure}

\subsection{Sim-to-Real Experiments}
To validate the practical utility of CarlaOcc, we conduct \textit{Sim-to-Real Transfer} experiments by pretraining models on CarlaOcc and finetuning on real-world datasets.
To bridge the domain gap, we map semantic labels and resample CarlaOcc occupancy GT to match the target dataset's semantic classes, voxel resolution, and spatial origin.
As shown in Table~\ref{tab:sim2real_results}, pretraining on CarlaOcc consistently improves mIoU by 0.8\%-1.5\% compared to training from scratch.
While seemingly modest, these gains are mainly constrained by the sparsity and noise of real-world GT (see Fig.~\ref{fig:comparison}, left), and by the configuration discrepancy, where KITTI-360 benefits more due to its closer setup to CarlaOcc.
Qualitatively (see Fig.~\ref{fig:comparison}), CarlaOcc pretraining yields more complete instance contours and improved occlusion handling, proving that CarlaOcc helps the model learn robust 3D spatial reasoning.

\subsection{Other 3D Perception tasks on CarlaOcc}
\noindent\textbf{Monocular Depth Estimation.}
Table~\ref{tab.depth} reports monocular depth estimation results on the CarlaOcc benchmark. We evaluate a representative set of state-of-the-art methods that cover three major paradigms: \emph{discriminative} models, \emph{generative} models, and \emph{metric} depth estimators. All methods take a single RGB image as input and are evaluated using standard depth metrics: absolute relative error (AbsRel), squared relative error (SqRel), root mean squared error (RMSE), log RMSE (RMSElog), and threshold accuracies~\cite{midas2023birkl}. Among discriminative approaches, DepthAnythingV2~\cite{depthanythingv2_2024yang} achieves the best overall performance, substantially improving over earlier networks such as MiDaS~\cite{midas2023birkl} and LeReS~\cite{leres2021yin} in both error and accuracy metrics. Generative methods such as Marigold~\cite{marigold2024ke} and LOTUS~\cite{lotus2024he} further reduce the AbsRel error, and the metric depth models (e.g., Metric3Dv2~\cite{metric3dv2_2024hu}, UniDepthV2~\cite{unidepthv2_2025piccinelli}) produce competitive or superior results while predicting depth in an absolute scale. These results indicate that CarlaOcc constitutes a challenging yet well-structured benchmark for advanced monocular depth estimators and can serve as a strong basis for downstream 3D perception tasks.

\begin{table}[!t]
	\centering
	\settablefont
	\setlength{\tabcolsep}{6pt}
	\caption{Representative 3D object detection baselines on the CarlaOcc dataset.}
	\label{tab.3d_det}
	\begin{tabular}{l|cccc}
		\toprule[1pt]
		Method 
		& mAP $\uparrow$
		& mATE $\downarrow$
		& mASE $\downarrow$
		& mAOE $\downarrow$ \\
		\midrule
		DETR3D~\cite{detr3d}
		& 0.39  & 0.63 & 0.26 & 0.37 \\
		PETR~\cite{petr}
		& 0.45  & 0.57 & 0.27 & 0.39 \\
		Sparse4D~\cite{sparse4d}
		& 0.48  & 0.66 & 0.24 & 0.33 \\
		\bottomrule[1pt]
	\end{tabular}
\end{table}

\noindent\textbf{3D Object Detection.}
Representative camera-only 3D object detection baselines are evaluated on our CarlaOcc benchmark. We follow the nuScenes evaluation protocol
and report mean Average Precision (mAP) together with the standard error metrics: translation error (mATE), scale error (mASE), and orientation error
(mAOE), where higher mAP and lower errors indicate better performance. We include three widely used multi-view detectors: DETR3D~\cite{detr3d}, PETR~\cite{petr}, and Sparse4D~\cite{sparse4d}. As shown in Table~\ref{tab.3d_det}, performance steadily improves from DETR3D to PETR and further to Sparse4D, with Sparse4D achieving the highest mAP among the tested methods. These baselines demonstrate that our dataset is compatible with state-of-the-art 3D detectors and can serve as a standard testbed for future research on camera-based 3D object detection.

{
	\small
	\bibliographystyle{ieeenat_fullname}
	\bibliography{main}
}

\end{document}